%% file: main.tex
\documentclass[runningheads]{llncs}

\usepackage{multirow}
\usepackage[mobile]{eccv}
\usepackage[pagebackref,breaklinks,colorlinks]{hyperref}
\usepackage{csquotes}

\hypersetup{colorlinks,linkcolor={blue},citecolor={blue},urlcolor={blue}}

\usepackage{eccvabbrv}
\usepackage{epsfig}
\usepackage{svg}

\usepackage{graphicx}
\usepackage{booktabs}

\usepackage[accsupp]{axessibility}  

\usepackage{pifont} 

\usepackage{comment} 

\makeatletter
\newcommand\footnoteref[1]{\protected@xdef\@thefnmark{\ref{#1}}\@footnotemark}
\makeatother

\usepackage[pagebackref,breaklinks,colorlinks]{hyperref}

\usepackage{orcidlink}

\begin{document}

\title{Meta-Prompting for Automating Zero-shot Visual Recognition with LLMs} 

\authorrunning{M. J. Mirza et al.}
\author{
\textbf{M. Jehanzeb Mirza\inst{1,2}}\quad
\textbf{Leonid Karlinsky\inst{3}}\quad
\textbf{Wei Lin\inst{4}}\\
\textbf{Sivan Doveh\inst{5,6}}\quad
\textbf{Jakub Micorek\inst{1}}\quad
\textbf{Mateusz Kozinski\inst{1}}\\
\textbf{Hilde Kuehne\inst{3,7}}\quad
\textbf{Horst Possegger\inst{1}}\\
{\institute{$^1$ICG, TU Graz, Austria. \quad $^2$CDL-EML. \quad $^3$MIT-IBM Watson AI Lab, USA. \quad $^4$JKU, Austria. \quad $^5$IBM Research, Israel. \quad $^6$Weizmann Institute of Science, Israel. \quad $^7$University of Bonn, Germany.}}
}

\maketitle
\input{def}
{\hspace{3em}Project Page: \href{https://jmiemirza.github.io/Meta-Prompting/}{https://jmiemirza.github.io/Meta-Prompting/}}
\input{sections/abstract}
\input{sections/introduction_mk}

\input{sections/related_work}

\input{sections/method_mk}

\input{sections/results}
\input{sections/conclusion}

\bibliographystyle{splncs04}
\bibliography{egbib}

\appendix
\clearpage
\section*{Supplementary Material}
\input{supp_sections/preamble}

\input{supp_sections/implementation_details}

\input{supp_sections/in_context_dataset}

\input{supp_sections/MMLM_pormpting_sensitivity}
\input{supp_sections/ensembling_descriptions}

\input{supp_sections/rebuttal_exp}
\input{supp_sections/detailed_results}
\end{document}

%% file: def.tex
\newcommand{\method}{MPVR\xspace}
\newcommand{\ClipEnc}{\theta}
\newcommand{\vis}{\phi}  
\newcommand{\txt}{\psi}  
\newcommand{\visEnc}{\vis} 
\newcommand{\txtEnc}{\txt} 

\newcommand{\txtEmb}{t} 
\newcommand{\txtEmbNew}{\txtEmb_{\hat{c}}} 

\newcommand{\imgEmb}{\vis} 

\newcommand{\cls}{c} 
\newcommand{\ClassNames}{C} 

\newcommand{\img}{x} 

\newcommand{\prompt}{p} 

\newcommand{\simil}{\cos} 

\newcommand{\likel}{l} 

\newcommand{\LSCE}{\mathcal{L}_\textsc{SCE}}

\newcommand\blue[1]{\textcolor{blue}{#1}}

\newif\ifdraft
\draftfalse
\drafttrue 
\ifdraft
 \newcommand{\MK}[1]{{\color{magenta}{\bf MK: #1}}}
  \newcommand{\JM}[1]{{\color{blue}{\bf JM: #1}}}
\newcommand{\LK}[1]{{\color{red}{\bf LK: #1}}}

 \newcommand{\mk}[1]{{\color{magenta} #1}}
\else
 \newcommand{\MK}[1]{}
 \newcommand{\mk}[1]{#1}
\fi

\newcommand{\quotebox}[1]{\begin{center}\fcolorbox{white}{blue!15!gray!15}{\begin{minipage}{1\linewidth}\vspace{1pt}\center\begin{minipage}{1\linewidth}{\space\Huge``}{#1}{\hspace{1.5em}\break\null\Huge\hfill''}\end{minipage}\smallbreak\end{minipage}}\end{center}}


 \newcommand{\tick}{\ding{51}}
  \newcommand{\cross}{\ding{55}}

  \definecolor{mycolor}{RGB}{176,64,35}

    \definecolor{llmcolor}{RGB}{255,18,18}
\newcommand{\flamingo}[1]{\textcolor{mycolor}{#1}}

\newcommand{\llmresponse}[1]{\textcolor{llmcolor}{#1}}

\newcommand\secvspace{\vspace{0cm}}
\newcommand\eqvspace{\vspace{0cm}}
\newcommand\figvspacetop{\vspace{0cm}}
\newcommand\figvspace{\vspace{0cm}}
\newcommand\tabvspace{\vspace{0cm}}
\newcommand\figcapvspace{\vspace{0cm}}

%% file: sections/abstract.tex
\begin{abstract}

Prompt ensembling of Large Language Model (LLM) generated category-specific prompts
has emerged as an effective method to enhance zero-shot recognition 
ability of
Vision-Language Models (VLMs).
To obtain these category-specific prompts, the present methods rely on hand-crafting the prompts to the LLMs for generating VLM prompts for the downstream tasks.
However, this requires manually composing these task-specific prompts and still, they might not cover the diverse set of visual concepts and task-specific styles associated with the categories of interest. 
To effectively take humans out of the loop and 
completely automate the prompt generation process for zero-shot recognition,
we propose \textbf{M}eta-\textbf{P}rompting for \textbf{V}isual \textbf{R}ecognition (\method).
%
Taking as input only minimal information about the target task, in the form of its short natural language description, 
and a list of associated class labels, 
\method automatically produces a diverse set of category-specific prompts resulting in a strong zero-shot classifier.
%
\method generalizes effectively across various popular zero-shot image recognition benchmarks belonging to widely different domains when tested with multiple LLMs and VLMs.
For example, \method obtains a zero-shot recognition improvement over CLIP by up to 19.8\% and 18.2\% (5.0\% and 4.5\% on average over $20$ datasets) leveraging GPT and Mixtral LLMs, respectively.
\end{abstract}

\vspace{-0.5cm}

%% file: sections/introduction_mk.tex
\secvspace
\section{Introduction}
\label{sec:introduction}
\secvspace

Dual encoder Vision-Language Models (VLMs)~\cite{clip,metaclip} attain unprecedented performance in zero-shot image classification.
They comprise a text encoder and an image encoder trained to map text and images to a shared embedding space. 
Zero-shot classification with dual encoder VLMs consists in evaluating the cosine similarity between the embedding of a test image and the embeddings of texts representing candidate classes.

The composition of the class-representing text has a significant impact on the accuracy of zero-shot classification.
Already the authors of CLIP~\cite{clip}, the first large-scale vision-language model, highlighted its importance and reported that embedding class names in a \emph{prompt} of the form \texttt{`A photo of a <class name>'} resulted in considerable performance growth over using raw class names.
Moreover, specializing the prompt to the data set by adding high-level concepts, for example, embedding the class name in the sentence \texttt{`A photo of a <class name>, a type of flower'} for fine-grained flower recognition, brought further improvement.
Finally, a substantial performance boost was achieved by ensembling multiple different prompts, tailored towards the downstream task (dataset). 
Since ensembling a larger number of dataset- and class-specific prompts is beneficial, and manually designing a large number of class-specific prompts is prohibitively time-consuming, several authors delegated prompt generation to a Large Language Model (LLM)~\cite{cupl,dclip,waffle}. 
These approaches consist in asking an LLM to generate class descriptions~\cite{cupl}, or class attributes~\cite{dclip}, and mix them with manually defined prompt templates~\cite{waffle}.
They enable generating large sets of prompts adapted to the downstream task, which would be prohibitively time-consuming when performed manually.
However, they
still require hand-crafting prompts to the LLM~\cite{cupl} or dataset-specific LLM prompt templates~\cite{waffle}, or rely on the assumption that class attributes are discriminative~\cite{dclip,waffle}.
In other words, they do not eliminate the manual effort completely, but shift some of it from manually designing prompts for the VLMs (as in~\cite{clip}) to manually designing LLM prompts.
This raises the following question:
Does the manual design of the LLM prompts bias the resulting VLM prompts, possibly affecting performance?
In this work, we answer this question affirmatively: we minimize manual interventions in the prompt generation process and show that this significantly boosts zero-shot recognition accuracy.
\footnote{To avoid confusion between the `prompts' used to query the LLMs and the `prompts' used to compute the text embedding by the VLMs, in the remaining part of this manuscript, we call the first one `LLM query' and the second one `VLM prompt'.}
\input{figs/motivation_fig}

The gist of our approach lies in automating the prompt generation process.
To that end, we draw inspiration from methods for reducing the prompt engineering effort in natural language processing~\cite{metaprompting,metaprompting1} 
and propose to meta-prompt the LLM to produce LLM query templates tailored to the downstream task.
We call our method Meta-Prompting for Visual Recognition (MPVR).
Its overview is presented in Figure~\ref{fig:motivation}.
MPVR comprises a `system prompt' that describes the meta-prompting task for the LLM, a description of the downstream task, and an in-context example.
The in-context example contains a description (metadata) of another task and its corresponding `LLM queries',
and serves to bootstrap the LLM with examples of expected results.
They are kept the same across different downstream tasks and for all our experiments.
MPVR extracts the LLM's knowledge of the visual world gradually, in two steps.
The first query to the LLM contains the system prompt, in-context example, and the downstream task (dataset) description, and produces a diverse set of LLM \emph{query templates}, containing a \texttt{<class name>} placeholder.
These templates are infused (by the LLM) with information on visual styles specific to the downstream task of interest, but they are still category-agnostic.
In the second step, for each class, we populate its label into all the task-specific LLM query templates generated in the first step and use them to query the LLM to generate (category-specific) VLM prompts describing the category in visually diverse ways and also containing task-specific visual styles infused by the LLM in the first step.
We use the resulting VLM prompts to create an ensemble of zero-shot classifiers.
In section~\ref{sec:results}, we show that MPVR's two-step process results in state-of-the-art zero-shot classification.

Our meta-prompting strategy does not take any parameters specific to the dataset, other than the dataset description, which can be easily obtained through public APIs or from its webpage. 
Yet, we show that prompts generated by MPVR cover diverse visual concepts and styles specific to the downstream task.
As a result, MPVR yields significant performance gains on a range of zero-shot benchmarks.
Our contributions can be summarized as follows: 
\begin{itemize}
    \item We propose MPVR: a general, automated framework requiring minimal human involvement for tapping into the visual world knowledge of LLMs through meta-prompting for zero-shot classification.
    \item MPVR generalizes beyond closed models (like GPT~\cite{gpt3}). We are the first to show that category-specific descriptions generated from open-source models (like Mixtral~\cite{mixtral}) can also enhance the zero-shot recognition abilities of state-of-the-art VLMs.
    \item We open-source a dataset of $\sim2.5\text{M}$ unique class descriptions harnessed from GPT and Mixtral with our meta-prompting framework. This is the first large-scale dataset encompassing the breadth of LLM knowledge of the visual world.
\end{itemize}

%% file: figs/motivation_fig.tex
\begin{figure*}[t!]
    \centering
    \includegraphics[width=0.85\textwidth]{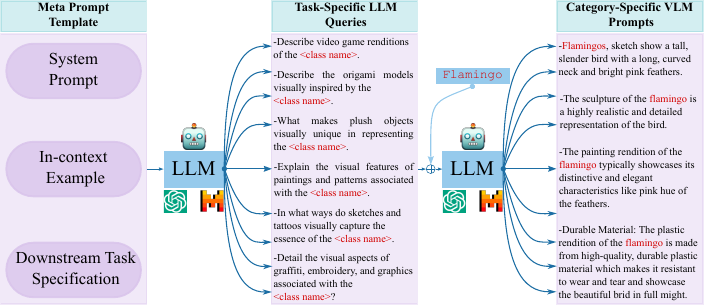}
    \caption{Our \method utilizes a Meta Prompt, comprising a system prompt (instruction), in-context example demonstrations (fixed throughout), and metadata (name and description) for a downstream task of interest. 
    The Meta Prompt instructs an LLM to generate diverse task-specific LLM queries, which are used to obtain category-specific VLM prompts (visual text descriptions) by again querying the LLM after specifying the \texttt{\color{red}{<class name>}}. These category-specific VLM prompts are then ensembled into a zero-shot classifier for recognizing the downstream task categories. 
    }
    \label{fig:motivation}
    \figvspace
\end{figure*}

%% file: sections/related_work.tex
\secvspace
\section{Related Work}
\label{sec:related-work}
\secvspace
We first provide an overview of the zero-shot vision-language foundation models, then touch upon approaches that propose to improve these models by requiring visual data (relying on additional training), and later discuss different methods that follow our line of work,~\ie improving zero-shot models in a training-free manner by generating textual data through LLMs and finally provide an overview of the prompt engineering literature. 

\secvspace
\paragraph{Large Scale Vision-Language Foundation Models:}

VLMs have shown impressive performance for many vision-language understanding tasks,~\eg zero-shot recognition, visual question-answering (VQA), image captioning,~\etc. 
The present-day VLMs can be placed in two distinct groups in a broader categorization. 
One group of methods relies on dual-encoders (vision and text encoder) and usually trains the encoders with a contrastive objective by using a large corpus of paired image-text data scraped from the web.
The most common among these methods are CLIP~\cite{clip}, ALIGN~\cite{align}, OpenCLIP~\cite{openclip}, and the very recent MetaCLIP~\cite{metaclip}.
The zero-shot classification is performed by measuring the similarity between the image embeddings and encoded text features, usually obtained by using the default template \texttt{`a photo of a <class name>'}.
The other group of methods aligns the visual modality with a frozen LLM. 
BLIP-2~\cite{blip2} bridges the modality gap between a pre-trained visual encoder and an LLM by using a querying transformer.
Instruct-BLIP~\cite{instructblip} proposes to improve~\cite{blip2} by employing instruction tuning. 
MiniGPT~\cite{minigpt} aligns a vision encoder with a frozen LLM (Vicuna~\cite{vicuna2023}) by only using a (trainable) linear projection layer between the two.  
MiniGPT-V2~\cite{minigptv2} replaces the LLM with Llama-2~\cite{touvron2023llama} and also proposes to unfreeze it during the training/finetuning phases. 
Llava~\cite{llava} also aligns an LLM with a pre-trained visual encoder and also proposes Visual Instruction Tuning, by carefully curating instruction-response pairs, to enhance the performance. 
Furthermore, the performance of LLaVA is also enhanced with better data curation~\cite{llava+} and slight architectural changes~\cite{llava-next}.
In our work, we focus on the contrastively pre-trained zero-shot models widely used for object recognition (\eg~CLIP~\cite{clip}), and improve the recognition abilities of these models by generating the text embeddings from a variety of descriptions (instead of the default templates) harnessed through our proposed meta-prompting technique. Furthermore, we show that \method-enhanced CLIP \cite{clip} outperforms even the leading LLM-decoder-based methods (\eg~\cite{llava-next}) in visual recognition tasks.

\secvspace
\paragraph{Training-based Approaches for Improving VLMs:}
Different approaches propose to improve the zero-shot recognition performance of the contrastively pre-trained models through parameter-efficient fine-tuning.
CoOp~\cite{coop} proposed to learn randomly initialized text prompts in a few-shot manner.
CoCoOp~\cite{cocoop} further conditions the learnable text prompts on the visual inputs to enhance the performance.  
Maple~\cite{maple} proposes to learn both the visual and text prompts in conjunction. 
Contrary to relying on few-shot labeled visual samples, UPL~\cite{upl} proposes to learn the text prompts on unlabeled image data and LaFTer~\cite{lafter} learns visual prompts by leveraging the cross-modal transfer capabilities of CLIP. 
While these approaches propose to adapt the VLM on image data, MAXI~\cite{maxi} proposes to fine-tune CLIP in an unsupervised manner for video inputs. 
In contrast to the methods proposed to improve the zero-shot recognition abilities of CLIP, our work does not rely on visual inputs and gradient-based updates of network parameters. 
Instead, it improves the zero-shot recognition performance by harnessing fine-grained textual concepts generated through our MPVR, thus supporting the capability to scale zero-shot recognition performance improvements to visual domains where \emph{no visual data} might be available for training. 

\secvspace
\paragraph{Zero-shot Recognition with Additional Textual Data from LLMs:} It was initially highlighted in CLIP~\cite{clip} that generating the text embeddings through an ensemble of (dataset specific) hand-crafted prompts\footnote{\href{https://github.com/openai/CLIP/blob/main/data/prompts.md}{https://github.com/openai/CLIP/blob/main/data/prompts.md}\label{footnote:prompt-ensemble-clip}} improved the zero-shot recognition performance on the downstream datasets, hinting towards the sensitivity of CLIP's text encoder towards fine-grained textual concepts.
Following up on this idea, DCLIP~\cite{dclip} enhances visual recognition by generating category-specific descriptors through an LLM (GPT-3~\cite{gpt3}).
On the other hand, CUPL~\cite{cupl} proposes to obtain the category-level text embeddings from the prompts generated with the dataset-specific hand-crafted queries fed to the LLM.
Waffle~\cite{waffle} hints towards the potential~\emph{bag-of-words} behavior of the CLIP text encoder and performs zero-shot classification by adding random descriptors to broad concepts and DCLIP-generated attributes.
Our work also takes inspiration from the prompt ensembling in~\cite{clip, cupl, dclip, waffle} and performs zero-shot classification by generating category-level prompts through an LLM. 
However, contrary to these approaches,~\method proposes a more general prompting framework to alleviate the human effort spent for handcrafting the LLM queries (CUPL~\cite{cupl}), dataset-specific concepts (Waffle~\cite{waffle}), or reduce reliance on individually recognizable visual attributes (DCLIP~\cite{dclip}). 
%
%
By effectively incorporating general downstream task information (description) into the first phase of \method (\ie~meta-prompting), we automatically produce task-tailored LLM query templates ready to be populated by task categories and used to query an LLM for a diverse spectrum of category-level VLM prompts comprising an enhanced set of visual details for recognizing those categories.
%
The performance gains by using \method with both closed and open-source LLMs (GPT~\cite{gpt3} and Mixtral~\cite{mixtral}) on $20$ different datasets when compared to relevant baselines highlight the generalization capabilities 
and benefits of our approach.  

\secvspace
\paragraph{Prompt Engineering:}

Manually manipulating the text inputs (prompts) to the LLMs for enhancing performance for various natural language processing (NLP) tasks has been an active field of research, which is formalized as prompt engineering.
In this context, providing demonstrations to the LLM for solving related downstream tasks has been referred to in the NLP literature as in-context learning (ICL)~\cite{icl, wei2021finetuned, chen2022improving, zhao2021calibrate}.
Orthogonal to the idea of in-context learning, some approaches rely on breaking down a complex task into a series of events. 
To this end, Chain-of-Thought (CoT)~\cite{cot} achieved impressive performance gains by prompting the model to perform intermediate reasoning steps.
Other approaches following this line of work include~\cite{kojima2022large, treeofthought}. 
Our \method also employs ICL and manipulates the input prompts to the LLMs, but effectively alleviates the need for human involvement for this manipulation by 
probing an LLM for more diverse concepts (LLM query templates -- incorporating general information about the task), which are then populated with specific task categories and fed again to the LLM for generating VLM prompts - both task- and category-specific text descriptions of visual concepts.
To the best of our knowledge, such a two-stage (meta-) prompting strategy for tapping into the visual world knowledge of LLMs does not exist in literature. 


%% file: sections/method_mk.tex
\secvspace
\section{MPVR: Meta-Prompting for Visual Recognition}
\label{sec:method}
\secvspace
Zero-shot classification with a dual encoder VLM consists in projecting a test image and each candidate class to the common embedding space,
and evaluating the cosine similarity between the embeddings.
The image embedding is produced by the VLM's vision encoder $\visEnc$.
The embedding of a class is obtained by passing a textual description of the class, called a VLM prompt, through the VLM's text encoder $\txtEnc$.
The simplest technique of constructing a VLM prompt is to complete a prompt template, for example,
`\texttt{A photo of a <class name>}', with class label~\cite{clip}.
The authors of CLIP~\cite{clip}, the first large-scale VLM, highlighted that prompt composition is vital to the performance of the zero-shot classifier.
To boost the performance, they proposed VLM prompt ensembling,
which represents the class as a mean embedding of multiple diverse prompts.
To formalize this approach, we denote the test image by $\img$, the set of candidate classes by $\ClassNames$, and the set of prompt templates by $P$. 
By $p(\cls)$ we denote a prompt obtained by completing template $p\in P$ with the label of class $\cls\in\ClassNames$. 
We define the zero-shot likelihood of class $\hat\cls$ as
\eqvspace
\begin{equation} \label{eq:mean_text_emb}
    \likel_{\hat\cls}(\img) = \frac{e^{\simil(\txtEnc_{\hat\cls},\visEnc(\img))/{\tau}}}{\sum_{\cls\in\ClassNames} e^{\simil(\txtEnc_{\cls},\visEnc(\img))/{\tau}}} ,
    \quad \text{where} \quad
    \txtEnc_{\cls} = \frac{1}{|P|} \sum_{p\in P} \txtEnc(p(\cls)),
\end{equation}
\eqvspace
and $\tau$ denotes the temperature constant.
This approach forms the point of departure for our method.

Ensembling a larger number of class-specific VLM prompts improves the performance of the zero-shot classifier, but generating these prompts manually would be prohibitively time-consuming.
Several methods~\cite{cupl,dclip,waffle,tap} address this problem by generating the VLM prompts with a large language model (LLM), for example GPT~\cite{gpt3}.
They enhance the performance of the zero-shot classifiers, but still require manual construction of the LLM queries,
which scales poorly: 
A prohibitively large human effort might be needed to creatively design prompts that cover the diverse ways the visual aspects of a certain class can be described in text.  
Moreover, manually specified queries can be influenced by the subjective bias of the person who composes them, which could affect zero-shot recognition performance.

To improve the scaling of VLM prompt generation and eliminate subjectivity from the process, we design Meta Prompting for Visual Recognition (MPVR), an approach to VLM prompt generation that reduces human input to the necessary minimum.
\method taps into the visual world knowledge possessed by the VLM and extracts it in two steps. 
In the first step, \method meta-prompts the LLM with generic instructions and coarse information about the downstream task to generate diverse task-specific LLM query templates. 
These LLM query templates encode elements of the LLM's knowledge about the visual styles characteristic of the downstream task but are still class-agnostic. 
In the second step, the LLM query templates are populated with names of candidate classes and fed to the LLM to obtain VLM prompts.
The resulting VLM prompts are both task- and class-specific.
Each prompt carries LLM's diverse visual knowledge about the possible appearance of objects representing the class in the style defined by the downstream task.

For ease of assimilation, we divide our \method into two distinct stages and provide an overview in Figure~\ref{fig:main}.
In Section~\ref{subsec:method:meta-prompting}, we describe how to effectively meta-prompt LLMs to generate diverse, task-specific LLM query templates (stage 1).
Later in Section~\ref{subsec:method:category-level-descriptions} we describe how to use these task-specific LLM query templates to obtain category-specific VLM prompts (stage 2).
\input{figs/method}


\secvspace
\subsection{Meta-Prompting a Large Language Model}
\label{subsec:method:meta-prompting}
\secvspace
Aligning with the true motivation of our \method, the goal of meta-prompting is to extract the abundant visual world knowledge possessed by the LLMs by querying it to generate multiple diverse LLM query templates with minimal human intervention. 
To that end, we compose a meta-prompt of three parts: the \textit{system prompt}, an \textit{in-context example}, and the \textit{downstream task specification}.
We illustrate the meta-prompt in Figure~\ref{fig:meta-prompting}.

\paragraph{System prompt} is a generic set of instructions that describe the elements of the meta-prompt and specify the expected output of the LLM. 
It instructs the LLM to generate a variety of query templates for the downstream dataset and conveniently format them to be employed in a~\texttt{Python} script.

\paragraph{In-context example} serves to bootstrap the LLM to the type of output that is expected.
It comprises a description of an example downstream task and a list of the corresponding LLM query templates.
Since we expect the output from the LLM to be suitable for use in a~\texttt{Python} script thus, it contains the prompts listed as~\texttt{Python code}~(\cf Figure~\ref{fig:meta-prompting}, middle left~\&~right).

\paragraph{Downstream task specification} is the only part of the meta-prompt that is specific to the downstream task. 
It is scraped from a public API or the webpage of the dataset associated with the task and contains a general description of the task data (\cf~Figure~\ref{fig:meta-prompting}, bottom left~\&~right). 
This coarse information about the downstream task of interest is critical for the LLM to generate task-specific LLM queries, which are employed in stage 2 of \method. 
\input{figs/meta_prompting_fig}

Note that the \textit{system prompt} and the \textit{in-context example} demonstrations are generic and are kept fixed across different tasks in all of our experiments.
The \textit{downstream task specification} is the only part of the meta-prompt that is specific to the downstream task. 
Our experiments highlight that all the individual parts of the meta-prompt are extremely vital for our \method to obtain effective category-specific VLM prompts and are extensively ablated in Table~\ref{tab:ablating-meta-prompt}.

The three elements of the meta-prompt are embedded in the template presented in Figure~\ref{fig:motivation} (left).
The resulting meta-prompt is then fed to the LLM (GPT~\cite{gpt3} or Mixtral~\cite{mixtral})
to generate $N$ diverse LLM query templates that are infused with the LLM's knowledge of visual styles expected in the dataset, but are still class-agnostic.
Instead of the downstream \texttt{<class name>} of interest, they contain a generic \texttt{<class name>} placeholder.
To obtain category-specific VLM prompts, we transition to stage $2$ of our~\method explained next.



\secvspace
\subsection{Category-Specific VLM prompts}
\label{subsec:method:category-level-descriptions}
\secvspace
The LLM response to the meta-prompt in stage 1 is a diverse set of LLM query templates, which contain task-specific knowledge about the downstream task of interest, but are still generic.
To instill the category information, for obtaining the category-specific VLM prompts,
we replace the generic \texttt{<class name>} placeholders in the LLM query templates with the actual class of interest.
These diverse category-specific queries constitute our second call to the LLM, which generates category-specific VLM prompts. 
They carry the LLM's knowledge of the appearance of objects of the queried classes in the context of the downstream task and are ready to be plugged into Eq.~\eqref{eq:mean_text_emb}.
We repeat this procedure for each class from 20 different datasets (used for evaluations) with both the GPT~\cite{gpt3} and Mixtral~\cite{mixtral} LLMs and obtain a huge corpus of $\sim2.5\text{M}$ VLM prompts. 
In section~\ref{sec:results}, we show that the ensemble of these VLM prompts results in a zero-shot classifier that outperforms previous methods by a significant margin.

The VLM prompts can be thought of as visually diverse descriptions of the queried classes in the context of the downstream tasks, and their corpus represents a chunk of the LLM's knowledge about our visual world.
This diversity stems from our proposed two-stage approach\footnote{We also experimented with generating category-specific VLM prompts in a single step with meta-prompting, but it performs worse than our 2-stage framework. These results are provided in the ablations Table~\ref{tab:abl:prompting-steps}.}.
The first stage can already provide diverse LLM query templates, which resemble the dataset-specific templates for prompt ensembling~\footnoteref{footnote:prompt-ensemble-clip} (but more diverse and automatically generated with our \method).
Interestingly, even by generating the ensemble of zero-shot classifiers by populating these generic query templates from stage 1 with category information, we can already achieve enhanced zero-shot recognition, as reported in an ablation in Table~\ref{tab:abl:prompting-steps}.
To conclude, after the second call to the LLM, the VLM prompts constitute fine-grained details about the specific category, reflecting the true diversity of the visual LLM knowledge and resulting in a huge category-specific text corpus,
already incorporated in our codebase released on this public Github repository: \href{https://github.com/jmiemirza/Meta-Prompting}{https://github.com/jmiemirza/Meta-Prompting}.

%% file: figs/method.tex
\begin{figure}[t!]
    \centering
    \includegraphics[width=0.8\textwidth]{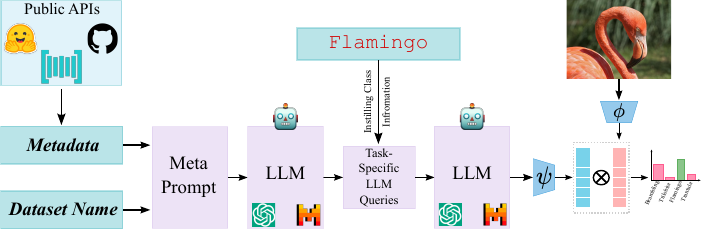}
    \caption{\method framework. In the first stage, a meta-prompt comprising of a \textit{system prompt}, \textit{in-context examples}, and metadata consisting of \textit{downstream task specification} is queried to the LLM instructing it to generate multiple diverse task-specific LLM queries, which are populated with the category of interest and again queried to the LLM 
    to obtain the category-level prompts for assembling a zero-shot classifier.}
    \label{fig:main}
    \figvspace
\end{figure}

%% file: figs/meta_prompting_fig.tex
\begin{figure*}[t!]
    \centering
    \includegraphics[width=1\textwidth]{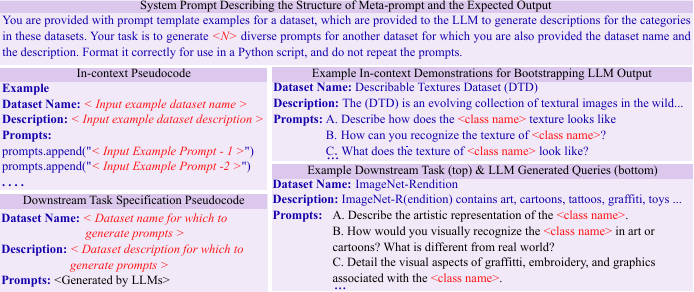}
    \caption{
    Our meta-prompt comprises $3$ parts: A \textit{system prompt} provides an overview of what is included in the overall prompt and what is expected from the LLM as a response (top). 
    An \textit{in-context example} consisting of metadata, dataset name, and hand-crafted prompts for the dataset (middle left). 
    The \textit{downstream task metadata} for which a diverse set of prompts are requested from the LLM (bottom left).
    For completeness, we also provide the in-context demonstrations (middle right) we use throughout, and the diverse LLM-generated queries for the example ImageNet-R dataset (bottom right). 
    }
    \label{fig:meta-prompting}
    \figvspace
\end{figure*}

%% file: sections/results.tex
\secvspace
\section{Experimental Evaluation}
\label{sec:results}
\secvspace
In this section, we first briefly describe the datasets and the baselines we use to evaluate and compare our \method, then explain our implementation details and finally provide a detailed discussion of the results.

\secvspace
\subsection{Evaluation Settings}
\secvspace
\paragraph{Datasets:} We extensively evaluate our \method on $20$ object recognition datasets belonging to widely different domains. 
These domains can be narrowed down to datasets containing commonly occurring natural categories: ImageNet~\cite{imagenet}, ImageNet-V2~\cite{imagenetv2}, CIFAR-10/100~\cite{cifar}, Caltech-101~\cite{caltech}.
Fine-grained classification datasets containing different task-specific images: Flowers~\cite{flowers102}, Standford Cars~\cite{cars}, CUBS-200~\cite{cubs200}, Oxford Pets~\cite{pets}, Describable Textures dataset (DTD)~\cite{dtd}, Food-101~\cite{food}, FGVC-Aircraft~\cite{aircraft}. 
Datasets used for scene classification: Places365~\cite{places} and SUN397~\cite{sun397}, action recognition datasets: UCF101~\cite{ucf101} and Kinetics400~\cite{k400}. 
Datasets consisting of out-of-distribution images: ImageNet-(R)endition~\cite{imagenet-r} and ImageNet-(S)ketch~\cite{imagenet-s} and also datasets which contain images taken from a satellite or an aerial view: EuroSAT~\cite{eurosat} and RESISC45~\cite{resisc}.

\secvspace
\paragraph{Baselines:}
We compare to the following baselines and state-of-the-art methods: 
\secvspace
\begin{itemize}
    \item \textbf{CLIP}~\cite{clip} denotes the zero-shot classification scores obtained by using the simple `\texttt{\{a photo of a <class name>\}}' template (S-TEMP) and dataset-specific templates (DS-TEMP\footnoteref{footnote:prompt-ensemble-clip}).

    \item \textbf{CUPL}~\cite{cupl} proposes to generate category-level descriptions from an LLM with hand-crafted prompts for each dataset. 

    \item \textbf{DCLIP}~\cite{dclip} proposes to obtain a zero-shot classifier with category-specific descriptors  (from an LLM) consisting of usual visual attributes.  
    \item \textbf{Waffle}~\cite{waffle} employs hand-crafted task-specific broad concepts and adds random descriptors to the prompts for zero-shot classification. 
    Following their evaluation setting, we compare with different variants: (i) Waffle (prompt + random descriptors), (ii) WaffleCon (Waffle + high-level concepts), and (iii) WaffleConGPT (WaffleCon + DCLIP descriptors). 
\end{itemize}

\input{tables/main_results}

\secvspace
\paragraph{Implementation Details:}
To report the results for each dataset we use the test splits provided by~\cite{coop} and further build upon their framework for all our evaluations on datasets that are not present in their framework.
All the baselines are also implemented in the same framework.
To generate the diverse set of task-specific LLM queries for our \method in the first stage, we use the public web API of ChatGPT\footnote{\href{https://chat.openai.com/}{https://chat.openai.com/}} and the Hugging Face API for Mixtral-7B (8x)\footnote{\href{https://huggingface.co/chat/}{https://huggingface.co/chat/}}.
To obtain the category-level VLM prompts after querying an LLM in the second stage of \method, we use GPT-3.5~\cite{gpt3} and the open source weights of Mixtral-7B (8x)~\cite{mixtral}, accessed through Hugging Face.
In the first stage, we instruct the LLM to generate $30$ diverse task-specific queries for each dataset, and to obtain the category-level VLM prompts, we append the category of interest and prompt the LLM to generate $10$ prompts for each LLM query respectively, where we limit each generated prompt by $50$ tokens.  
The in-context dataset is (arbitrarily) chosen to be DTD~\cite{dtd} for all experiments, however, to avoid information contamination, we switch the in-context dataset to EuroSat~\cite{eurosat} when DTD is the target dataset.
We ablate the choice of DTD for in-context example and provide the complete meta prompts in the supplementary. 
Unless otherwise specified, we obtain the zero-shot classifier as the mean of the class embeddings obtained from the category-specific VLM prompts (from stage 2 of MPVR) using Eq.~\eqref{eq:mean_text_emb}. 

\secvspace
\subsection{Results}
\label{subsec:results}
\secvspace
We test our \method extensively on $20$ diverse datasets and report the results (with ViT-B/32 from CLIP~\cite{clip}) 
in Table~\ref{tab:main-results}.
We consistently outperform the CLIP zero-shot baseline while using the category-level prompts generated both from GPT and Mixtral. 
While comparing to the CLIP baseline, using the default template, on some datasets like EuroSAT, the improvement is up to 19.8\% and 18.2\%, and on average our \method improves upon CLIP by 5.0\% and 4.5\% while averaging the results on $20$ datasets, with GPT and Mixtral LLMs respectively.
Similarly, while compared to the more expressive CLIP zero-shot baseline, which uses the hand-crafted dataset-specific templates\footnoteref{footnote:prompt-ensemble-clip}, we still observe considerable average gains of 3.1\% and 2.7\% with the two LLMs. 

Our \method also shows strong gains when compared to CUPL~\cite{cupl}, which obtains category-level prompts by hand-crafting the LLM queries for each downstream task of interest. 
Our \method 
not only alleviates this extensive human effort spent while generating the category-level prompts (as in CUPL~\cite{cupl}) but also out-performs CUPL on most of the datasets we compare to. 
For example, obtaining up to $5.1\%$ and $6.3\%$ performance gains on Flowers-102~\cite{flowers102} dataset with GPT and Mixtral LLMs.
\input{tables/different_backbones}

Furthermore, we also observe that while comparing with the baselines which do not generate (descriptive) VLM prompts but rely on other cues like category-level (attribute) descriptors, our \method also performs favorably. 
For example, we outperform DCLIP~\cite{dclip} on all the $20$ datasets with performance gains up to $5.3\%$ and $3.3\%$ on UCF-101 with GPT and Mixtral. 
These results indicate that the generic attributes generated for a category by DCLIP for classification might not capture fine-grained task-specific details required to enhance the classification of categories in domain-specific benchmarks (\eg~action recognition in UCF-101).
Finally, from Table~\ref{tab:main-results} we also observe that our \method (on average) also outperforms all the variants proposed by Waffle~\cite{waffle}, which also highlights that the CLIP text encoder responds favorably to semantically rich text descriptions (prompts), instead of randomly generated descriptors as in Waffle~\cite{waffle}.
In summary, our \method demonstrates better performance across the board, outperforming all baselines on 18 out of 20 datasets. 
On the Food-101~\cite{food} dataset, our \method comes in second, trailing by a narrow margin of 0.3\%. 
Similarly, on Standford Cars~\cite{cars}, our results indicate that even the dataset-specific prompt ensembling proposed by CLIP fails to enhance performance, underscoring the unique challenges posed by this particular dataset.

To test the generalization ability of our \method beyond different LLMs, we also evaluate it with different backbones from CLIP~\cite{clip} and also employ MetaCLIP~\cite{metaclip}, which is trained with a different training recipe than CLIP. 
These results are listed in Table~\ref{tab:main-res:different-backbones}. 
We observe that even while testing with more expressive backbones, like MetaCLIP ViT-L/14, our visually diverse text descriptions (prompts) help to improve the zero-shot accuracy from $71.0\% \to 74.3\%$ (for GPT descriptions) while averaging over the $20$ datasets.    
Due to space limitations, we defer the individual dataset results for these backbones to the supplementary. 

\secvspace
\subsection{Ablations}
\label{subsec:results:ablations}
\secvspace
Here, we study the significance of different components that constitute our \method. 
Specifically, we first examine the effect of combining multiple text sources, and then 
motivate our choice of using dual encoder models like CLIP~\cite{clip} instead of multi-modal language models (MMLMs) by evaluating them for image classification.
Later we extensively ablate our prompting strategy and finally conclude with ablations on robustness of the obtained results and scaling analysis.

\input{tables/mixture_of_descriptions}
\input{tables/llava_results}

\secvspace
\paragraph{Ensembling Text Sources.}
From Tables~\ref{tab:main-results}~\&~\ref{tab:main-res:different-backbones} we gather that in addition to the enhanced zero-shot classification with GPT and Mixtral generated VLM prompts with our \method, the dataset-specific templates\footnoteref{footnote:prompt-ensemble-clip} from CLIP can also show improvement in results, in comparison to only using the default templates. 
To evaluate the combined performance of these text sources, we ensemble the $3$ different sources and provide the results in Table~\ref{tab:mixture-of-ensembles}.
We observe that when the category-specific VLM prompts and templates are ensembled over the embedding space, the resulting classifier is weaker than the classifier obtained from only the LLM-generated VLM prompts. 
However, the mean of the embeddings from both GPT and Mixtral prompts performs the best. 
These results hint that the prompts from both the LLMs are clustered closely in the CLIP latent space suggesting that these sources describe the categories of interest in a similar (more detailed) way, yet differently from the `more mechanical' CLIP dataset-specific prompts that do not provide much detail.
We also test ensembling the probability spaces from both sources and find that the degradation in performance as a consequence of mixing the descriptions and templates is alleviated. 
%
%

\secvspace
\paragraph{MMLMs for Zero-shot Classification.}
Recently, multi-modal language models such as LLaVA~\cite{llava, llava+} have emerged as the preferred choice for various vision-language tasks. 
Here, we extended their evaluation to zero-shot classification, and the findings are summarized in Table~\ref{tab:llava1.6-classification-results}. 
Notably, our results indicate that, for the specific task of object recognition, CLIP~\cite{clip} outperforms LLaVA by a substantial margin, reinforcing our decision to employ CLIP for the discriminative task, which is the focus of our study. 
We ablate and detail the sensitivity of MMLMs to different prompting strategies in the supplementary, here we report only its best prompting strategy result.
\input{tables/ablating_meta_prompting}
\input{tables/abl_eurosat_1_step_prompts_only}

\secvspace
\paragraph{Meta Prompt.}
In Table~\ref{tab:ablating-meta-prompt} we ablate different components of our meta-prompt (outlined in Figure~\ref{fig:meta-prompting}) and report the results on the EuroSAT dataset. 
We see that all the major components have a strong effect on the downstream performance.
For example, if we do not populate the meta-prompt with the in-context demonstrations of example LLM queries for a dataset, the LLM fails to generate the task-specific queries from the first stage. 
Similarly, removing the metadata (description of datasets) from the in-context example and the resulting dataset of interest also results in a huge performance drop $55.6\%\to42.0\%$. 
We also noticed that interestingly, providing the category names for the datasets in the meta prompt (for stage 1) as extra information did not improve the results, potentially hinting that LLM prefers more simple and succinct instructions.

\secvspace
\paragraph{Altering Meta Prompting Stages.}
In Table~\ref{tab:abl:prompting-steps} we report the results by altering our meta-prompting strategy in two distinct ways: 
By generating the category-level VLM prompts directly in one step, by incorporating the class name already in stage 1 of our \method, and populating the \texttt{<class names>} in the generated task-specific LLM queries from stage 1 (which resembles the prompt ensembling performed by CLIP~\cite{clip}).
The results indicate that our 2-stage approach performs better than altering it to a single stage, and even our generated prompts from stage 1 can offer a more robust zero-shot classifier than templates ensembling\footnoteref{footnote:prompt-ensemble-clip}, highlighting the visual diversity of our generated task-specific queries, which later effectively translates to the VLM prompts as well. 
\input{figs/scaling_std_tab}

\secvspace
\paragraph{Results Robustness and Scaling Analysis: }
In Table~\ref{tab:robust} we study the robustness of \method results by reporting the mean and variance with randomly sampling \method-generated VLM prompts $10$ times for all $20$ datasets. 
We observe that the variances are negligible \wrt the obtained gains (in Table~\ref{tab:main-results}).
In Figure~\ref{fig:scaling} we show the scaling potential by sampling more VLM category- and task-specific prompts. 
The results highlight that sampling an increasing number of generated VLM prompts significantly boosts performance showing promising scaling potential.





%% file: tables/main_results.tex
\begin{table*}[t!]
    \centering

\resizebox{\textwidth}{!}{\begin{tabular}{lcccccccccc}
\toprule
 & \textbf{ImageNet} & \textbf{ImageNetv2} & \textbf{C10} & \textbf{C100} & \textbf{Caltech101} & \textbf{Flowers} & \textbf{Stanford Cars} & \textbf{Cubs} & \textbf{Pets} & \textbf{DTD} \\
\midrule
\midrule
CLIP (S-TEMP) & {61.9} & {54.8} & {88.3} & {64.4} & {91.4} & {64.0} & {\bfseries{60.2}} & {51.6} & {85.0} & {40.2} \\
CLIP (DS-TEMP) & {63.3} & {56.0} & {89.2} & {65.1} & {89.9} & {66.7} & {\underline{60.0}} & {53.0} & {87.4} & {42.4} \\
CUPL & {\underline{64.3}} & {\underline{56.9}} & {89.0} & {65.3} & {92.1} & {68.8} & {\underline{60.0}} & {51.9} & {87.2} & {48.9} \\
DCLIP & {63.1} & {55.8} & {86.7} & {64.2} & {92.5} & {64.6} & {57.9} & {52.6} & {83.5} & {44.3} \\
Waffle & {63.4} & {56.3} & {89.4} & {65.2} & {90.8} & {67.8} & {59.9} & {52.8} & {87.7} & {40.4} \\
Waffle+Con & {63.4} & {56.3} & {89.4} & {65.2} & {89.7} & {65.2} & {59.5} & {52.1} & {86.8} & {41.7} \\
Waffle+Con+GPT & {63.4} & {56.3} & {89.4} & {65.2} & {91.9} & {68.2} & {59.6} & {52.6} & {87.9} & {41.8} \\
\midrule
\method (MIXTRAL) & {63.8} & {56.5} & {\underline{89.5}} & {\underline{65.5}} & {\underline{92.8}} & {\bfseries{75.2}} & {58.3} & {\underline{55.5}} & {\underline{88.0}} & {\underline{50.2}} \\
\method (GPT) & {\bfseries{65.0}} & {\bfseries{57.3}} & {\bfseries{89.9}} & {\bfseries{66.3}} & {\bfseries{92.9}} & {\underline{73.9}} & {59.5} & {\bfseries{55.9}} & {\bfseries{88.1}} & {\bfseries{50.8}} \\
\midrule
& \textbf{Food101} & \textbf{Aircraft} & \textbf{Places365} & \textbf{SUN397} & \textbf{UCF101} & \textbf{K400} & \textbf{IN-R} & \textbf{IN-S} & \textbf{EuroSAT} & \textbf{Resisc45} \\
\midrule
\midrule
CLIP (S-TEMP) & {77.6} & {18.1} & {39.4} & {62.1} & {60.4} & {39.7} & {66.3} & {41.1} & {35.9} & {54.1} \\
CLIP (DS-TEMP) & {79.2} & {19.5} & {40.0} & {63.0} & {62.4} & {42.1} & {69.3} & {42.7} & {45.8} & {57.8} \\
CUPL & {81.0} & {20.4} & {\_} & {\underline{66.5}} & {65.2} & {41.7} & {\_} & {\_} & {\_} & {61.9} \\
DCLIP & {79.7} & {19.8} & {40.9} & {63.1} & {62.6} & {39.1} & {66.0} & {42.3} & {48.9} & {56.9} \\
Waffle & {\bfseries{81.6}} & {20.1} & {41.1} & {63.3} & {62.7} & {40.4} & {68.8} & {43.4} & {42.7} & {61.4} \\
Waffle+Con & {81.1} & {19.0} & {39.3} & {60.7} & {62.2} & {39.1} & {68.1} & {42.5} & {44.8} & {58.6} \\
Waffle+Con+GPT & {81.2} & {19.8} & {41.5} & {64.0} & {63.4} & {40.4} & {68.5} & {\underline{43.7}} & {47.0} & {62.0} \\
\midrule
\method (MIXTRAL) & {\underline{81.3}} & {\bfseries{22.4}} & {\underline{42.1}} & {\underline{66.5}} & {\underline{66.0}} & {\underline{42.2}} & {\bfseries{70.2}} & {43.6} & {\underline{54.0}} & {\bfseries{64.6}} \\
\method (GPT) & {81.0} & {\underline{21.5}} & {\bfseries{42.2}} & {\bfseries{67.0}} & {\bfseries{67.9}} & {\bfseries{43.9}} & {\bfseries{70.2}} & {\bfseries{44.2}} & {\bfseries{55.6}} & {\underline{64.0}} \\
\bottomrule
\bottomrule
\end{tabular}}
\caption{Top-1 accuracy (\%) for $20$ datasets obtained by employing the ViT-B/32 backbone from OpenAI CLIP~\cite{clip}. \emph{S-TEMP} refer to the results obtained by using the default template (\texttt{a photo of a <class name>}), while \emph{DS-TEMP} refer to the results obtained by using the ensemble of dataset specific prompts.
An empty placeholder for CUPL~\cite{cupl} indicates that the respective baseline did not provide the handcrafted prompts for the dataset. 
For Waffle~\cite{waffle}, mean results from $7$ random runs are reported, following the original publication.
}
\label{tab:main-results}
\tabvspace
\vspace{-0.4cm}
\end{table*}

%% file: tables/different_backbones.tex

\begin{table*}[t!]
    \centering
        \setlength{\tabcolsep}{10pt}

    \resizebox{0.7\textwidth}{!}{\begin{tabular}{l|cc|cccc}
    \toprule
    &\multicolumn{2}{c|}{OpenAI CLIP}&\multicolumn{3}{c}{MetaCLIP 400m}\\
    \cmidrule{2-3} \cmidrule{4-6}
    &  B/16 & L/14 & B/32&B/16&L/14 \\
    \midrule
    \midrule
S-TEMP & {61.9} & {69.2} & {62.4} & {65.9} & {71.0} \\
DS-TEMP & {63.8} & {71.2} & {64.0} & {67.3} & {72.8} \\
D-CLIP & {64.4} & {70.7} & {62.8} & {66.4} & {72.2} \\
Waffle & {64.0} & {70.7} & {62.8} & {66.5} & {72.4} \\
Waffle+Con & {62.7} & {69.1} & {61.7} & {65.7} & {71.7} \\
Waffle+Con+GPT & {64.6} & {71.0} & {63.2} & {66.9} & {72.7} \\
\midrule
\method (Mixtral) & {\underline{66.4}} & {\underline{72.5}} & {\underline{65.6}} & {\bfseries{68.7}} & {\underline{73.9}} \\
\method (GPT) & {\bfseries{66.7}} & {\bfseries{73.4}} & {\bfseries{65.8}} & {\bfseries{68.7}} & {\bfseries{74.3}} \\
\bottomrule
\bottomrule
    \end{tabular}}
    \caption{Mean top-1 accuracy (\%) over $20$ datasets for different backbones from OpenAI~\cite{clip} and MetaCLIP-400m~\cite{metaclip}.}
    \label{tab:main-res:different-backbones}
    \tabvspace
    \vspace{-0.4cm}
\end{table*}

%% file: tables/mixture_of_descriptions.tex
\begin{table*}[t!]
\small
\centering
\setlength{\tabcolsep}{2pt}
\resizebox{1\textwidth}{!}{
\begin{tabular}{ccc|ccc|cccc}
\toprule
& && GPT & Mixtral & Temp& GPT+Temp & Mixtral+Temp & GPT+Mixtral & GPT+Mixtral+Temp \\
\midrule
\midrule
\multirow{3}{*}{\rotatebox[origin=c]{90}{\tiny{Embedding}}} & \multirow{3}{*}{\rotatebox[origin=c]{90}{\tiny{Average}}}&ViT-B/32 & \text{62.9} & 62.4 & 59.7&57.0 & 56.1 & \text{63.0} & 57.7 \\
&& ViT-B/16 & \text{66.7} & 66.4 &63.8 & 60.5 & 59.6 & \text{67.0} & 61.5 \\
&& ViT-L/14 & \text{73.4} & \text{72.5}& 71.2& 68.6 & 67.3 & \text{73.4} & 69.2 \\
\midrule
\multirow{3}{*}{\rotatebox[origin=c]{90}{\tiny{Softmax}}} &\multirow{3}{*}{\rotatebox[origin=c]{90}{\tiny{Average}}}& ViT-B/32 & \_ & \_ &59.8 & \text{62.8} & 62.3 & 62.4 & 62.4 \\
&& ViT-B/16 & \_ & \_ & 63.8&\text{66.7} & \text{66.4} & \text{66.4} & 66.3 \\
&& ViT-L/14 & \_ & \_ &71.1 &\text{73.3} & 72.4 & 72.6 & 72.6 \\
\bottomrule
\bottomrule
\end{tabular}
}
\caption{Comparison of mean top-1 accuracy (\%) for \method over 20 datasets while constructing the zero-shot classifier by ensembling with the mean of the embeddings from different text sources (top) and mean of softmax (bottom). For GPT and Mixtral, we only report the results with the mean of the embeddings, since ensembling the softmax of individual descriptions is prohibitively expensive (also noted in~\cite{clip}). 
For datasets with fewer classes, we performed softmax ensembling but did not find any major deviation in results. These results are provided in the supplementary. }
\label{tab:mixture-of-ensembles}
\tabvspace
\end{table*}


%% file: tables/llava_results.tex
\begin{table*}[t!]
    \centering
            \setlength{\tabcolsep}{4pt}
    \resizebox{0.7\textwidth}{!}{\begin{tabular}{l|ccccc|c}
    \toprule
         &  EuroSAT & DTD &Caltech&CIFAR-100&Resisc&Mean\\
                 \midrule
        \midrule
        CLIP (ViT-B/32)& 35.9 &40.2 &91.4&64.4&54.1&57.2\\

        LLAVA-1.6 (7B)& 41.3 &16.2 &33.0&25.7&33.8&30.0 \\
        \midrule
         MPVR (ViT-B/32)& 55.6&50.8&92.9&66.3&64.0&65.5\\
         \bottomrule
         \bottomrule
         
    \end{tabular}}
    \caption{Comparison of top-1 accuracy (\%) with LLAVA-1.6-Vicuna7b model~\cite{llava+}.}
    \label{tab:llava1.6-classification-results}
    \tabvspace
    \vspace{-0.4cm}
\end{table*}

%% file: tables/ablating_meta_prompting.tex


         

\begin{table*}[t!]
\small
    \centering
        \setlength{\tabcolsep}{4pt}
    \resizebox{0.7\textwidth}{!}{\begin{tabular}{cccc|c}
    \toprule    
 dataset name & dataset metadata &in-context (prompts) & class names&Top-1\\
\midrule
\midrule
\cross & \tick & \tick&\cross&46.7\\

\tick & \cross & \tick&\cross&42.0\\

\tick & \tick & \cross&\cross&\_\\

\tick & \tick & \tick&\tick&53.5\\

\midrule

\tick & \tick & \tick&\cross&55.6\\


         \bottomrule
         \bottomrule
         
    \end{tabular}}
    \caption{Top-1 accuracy (\%) for EuroSAT~\cite{eurosat} with GPT as LLM and the ViT-B/16 backbone~\cite{clip} while ablating the different parts of our Meta Prompt. The last row represents the results obtained by our \method.}
    \label{tab:ablating-meta-prompt}
    \tabvspace
\end{table*}


    

%% file: tables/abl_eurosat_1_step_prompts_only.tex
\begin{table*}[t!]
    
                \setlength{\tabcolsep}{4pt}

    \centering
    
    \resizebox{0.7\textwidth}{!}{\begin{tabular}{c|ccccc}
    \toprule
         & CLIP (S-TEMP) &CLIP (DS-TEMP)  & Prompts-Only& 1-Step& MPVR\\
         \midrule\midrule
        EuroSAT & 35.9 &45.8&47.2&51.2&55.6\\
        \bottomrule
        \bottomrule
    \end{tabular}}
    \caption{Comparison of top-1 accuracy~(\%) from the zero-shot classifier obtained with the prompts generated in the first stage and generating category-level descriptions directly from stage-1 of \method.}
    \label{tab:abl:prompting-steps}
    \tabvspace
    \vspace{-0.4cm}
\end{table*}

%% file: figs/scaling_std_tab.tex
\begin{figure*}[t!]
\vspace{2em}
\figvspace
\begin{minipage}{0.45\textwidth}
\centering
\begin{tabular}{c|cc}
    \toprule
                    & accuracy(\%)& std\\
                    \midrule
                    \midrule
         ViT-B/32& 62.8 & $\pm$0.05\\
         ViT-B/16& 66.7 & $\pm$0.04\\
         ViT-L/14& 73.3 & $\pm$0.03\\
         \bottomrule
         \bottomrule
    \end{tabular}
    \captionsetup{type=table}
    \vspace{0.4em}
    \caption{Top-1 mean accuracy (\%) for CLIP and standard deviation for $10$ random runs, for all datasets.}\label{tab:robust}
\end{minipage}
\hfill
\begin{minipage}{0.5\textwidth}
    \vspace{-1.2em}
\includegraphics[width=1\textwidth]{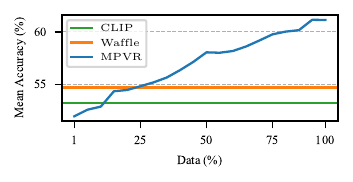}
\centering
    \vspace{-3.1em}
    \caption{Top-1 mean accuracy (\%) over DTD, EuroSat, Flowers, Resisc45, subsampling the VLM prompts sets.}\label{fig:scaling}
\end{minipage}
\figvspace
\end{figure*}

%% file: sections/conclusion.tex
\secvspace
\section{Conclusion}
\secvspace

We have presented meta-prompting for enhancing zero-shot visual recognition with LLMs, which essentially alleviates any human involvement in VLM prompt design for new tasks.
Our \method generates task-specific category-level VLM prompts by only requiring minimal information about the downstream task of interest.
\method first queries the LLM to generate different high-level queries letting it discover the diverse ways of querying itself to generate visually diverse category-level prompts.
These prompts are ensembled to construct a robust zero-shot classifier, that achieves enhanced zero-shot classification on a diverse set of $20$ datasets belonging to widely different domains.  
Furthermore, we also open-source the $2.5\text{M}$ category-level text descriptions dataset, harnessed from GPT and Mixtral, covering the breadth of the LLM knowledge of our visual world.
This large-scale dataset can be employed in many exciting future work directions,~\eg fine-tuning multi-modal language models for enhanced fine-grained visual classification, or constructing large-scale synthetic datasets via generative text-to-image models for VLM pre-training.  

%% file: supp_sections/preamble.tex
As supplementary material for our \method: Meta Prompting for Visual Recognition, we first 
list additional implementation details (Section~\ref{sec:supp:implementation_details}).
Then, for additional insights, we provide an ablation on the use of the in-context dataset employed for meta-prompting (Section~\ref{sec:supp:in_context_dataset}). 
Moving forward, we provide results with different strategies employed for prompting multimodal language models (MMLMs) for the task of object recognition (Section~\ref{sec:supp:prompt_eng_for_mmlms}), demonstrating we used the best performing available strategy for the MMLM baseline in the main paper. 
Then, we provide results for ensembling (in probability space) the vision language model (VLM) prompts generated through our MPVR (Section~\ref{sec:supp:ensembling_descriptions}). 
Later, 
we conclude with experiments performed during the rebuttal phase (Section~\ref{sec:supp:additional-insights-experiments}) and detailed (dataset-wise) results (Section~\ref{sec:supp:detailed_results}). 


%% file: supp_sections/implementation_details.tex
\section{Implementation Details}
\label{sec:supp:implementation_details}
All our experiments are performed on a single NVIDIA 3090 GPU. 
To obtain the results for the baselines, we use their official codebase and run the baselines locally with all their recommended parameters and settings. 
For CUPL~\cite{cupl} we only report the results on the datasets, for which the authors provided the category-level VLM prompts. 
Since CUPL uses hand-crafted dataset-specific LLM queries to generate the category-level VLM prompts, for some datasets these queries are not available, so we were not able to generate the VLM prompts for those datasets. 
We used the category-level VLM attributes provided by DCLIP~\cite{dclip} in their official repository\footnote{\href{https://github.com/sachit-menon/classify_by_description_release}{https://github.com/sachit-menon/classifybydescriptionrelease}}. 
For the datasets, not listed in their repository, we used their official code to generate the attributes and used them for obtaining the Waffle~\cite{waffle} results, following the official publication.
In contrast to CUPL~\cite{cupl}, the attributes can be generated for any dataset, only by providing the class names from the downstream datasets. 
Similarly, following the official publication and settings proposed in Waffle~\cite{waffle}, the datasets for which the high-level concepts are not available (\ie~ImageNet~\cite{imagenet}, ImageNetv2~\cite{imagenetv2}, CIFAR10/100~\cite{cifar}), their two variants, Waffle+Con and Waffle+Con+GPT, collapse to only the Waffle results, in all the tables.

%% file: supp_sections/in_context_dataset.tex
\section{Meta Prompt}
\label{sec:supp:in_context_dataset}

In the main manuscript, we arbitrarily employed the Describable Textures Dataset (DTD)~\cite{dtd} as the in-context example dataset for all our experiments. 
However, when the target dataset is DTD, we switched the in-context example dataset to EuroSAT~\cite{eurosat}.
Here, we studied the effect of employing different in-context datasets.
For example, when employing an alternative in-context dataset, such as Flowers~\cite{flowers102} or CUBS~\cite{cubs200} for DTD (as the target dataset), the variance in results is only $\pm 0.71$, considerably lower than the gains of $8.4\%$ ($50.8\%$ \textit{vs.} $42.4\%$) obtained over the baseline of CLIP + `dataset-specific templates', for the ViT-B/32 backbone from CLIP~\cite{clip}. 

Similarly, while using an alternative in-context dataset, Flowers or Cubs, for the target dataset EuroSAT, the variance in obtained results is only $\pm 0.44$, again considerably lower than the gains of $9.8\%$ ($55.6\%$ \textit{vs.} $45.8\%$) obtained over the baseline of CLIP + `dataset-specific templates'. 
\input{supp_figs/meta_prompt}
Furthermore, for completeness, we also provide $2$ complete meta-prompt examples in Figure~\ref{fig:supp:meta-prompt} while choosing different in-context demonstrators (\ie~DTD~\cite{dtd} and Flowers~\cite{flowers102}) and target datasets (\ie~ImageNet-R~\cite{imagenet-r} and DTD~\cite{dtd}).

%% file: supp_figs/meta_prompt.tex
\begin{figure}[t!]
    \centering
    \includegraphics[width=\textwidth]{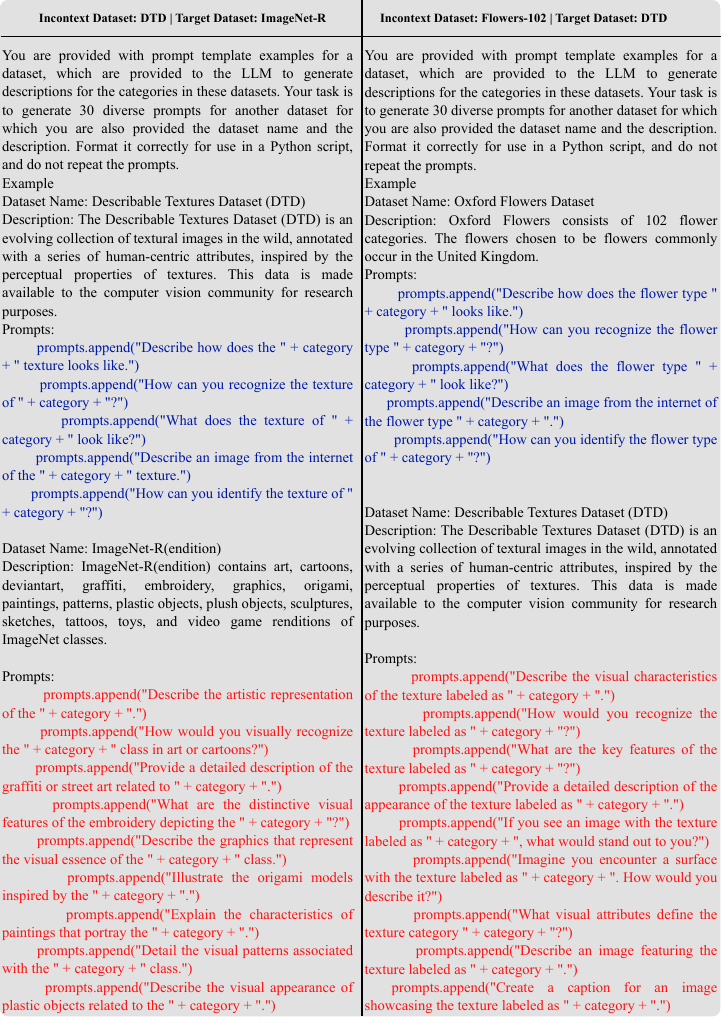}
    \caption{Exemplary meta-prompts (and a few \llmresponse{LLM generated responses}) for \method using different in-context (left: DTD~\cite{dtd}, right: Flowers~\cite{flowers102}) and target (left: ImageNet-R~\cite{imagenet-r}, right: DTD~\cite{dtd}) datasets.}
    \label{fig:supp:meta-prompt}
\end{figure}

%% file: supp_sections/MMLM_pormpting_sensitivity.tex
\section{Prompt Engineering for MMLM}
\label{sec:supp:prompt_eng_for_mmlms}
To address the sensitivity of MMLMs to different prompting strategies, we extensively tested the following different prompting variations used for the task of category recognition for MMLMs. These prompting strategies are also illustrated in Figure~\ref{fig:llava_prompting} for the EuroSAT dataset.

\paragraph{Categories as Numbered Options:} The prompt to the MMLM~\cite{llava-next} contained the categories (the model needed to choose from) listed as numbered options. 
\paragraph{Categories as Alphabet Options:} The prompt to the MMLM~\cite{llava-next} contained the categories (the model needed to choose from) listed as English alphabet options. 
\paragraph{Categories as List:} In this prompting strategy, we provided the category names as a list and the MMLM was prompted to output the exact name of the category for each test image. 

In Table~\ref{tab:ablating-prompting-for-llava} we list the results for different prompting strategies and find that the best results were obtained
when LLAVA-1.6~\cite{llava-next} was prompted with categories (to choose from) as numbered options.
For the fairest comparison, the LLAVA-1.6~\cite{llava-next} baseline results reported in Table $4$ of the main manuscript were obtained using this (top-performing) prompting option for all the tested datasets.
\input{tables/diff_llava_evalations}
\input{supp_figs/llava_prompting}

%% file: tables/diff_llava_evalations.tex
\begin{table*}[t!]
    \centering
            \setlength{\tabcolsep}{8pt}
    \begin{tabular}{c|ccc}
    \toprule
         &  Numbered Options & Alphabet Options & List Option\\
         \midrule
         \midrule
         EuroSAT& 41.3 &38.7&34.4\\ 
         \bottomrule
         \bottomrule
    \end{tabular}
    \caption{Top-1 accuracy~(\%) with different prompting strategies for LLAVA-1.6~\cite{llava+}.}
    \label{tab:ablating-prompting-for-llava}
\end{table*}

%% file: supp_figs/llava_prompting.tex
\begin{figure}
    \centering
    \includegraphics[width=\textwidth]{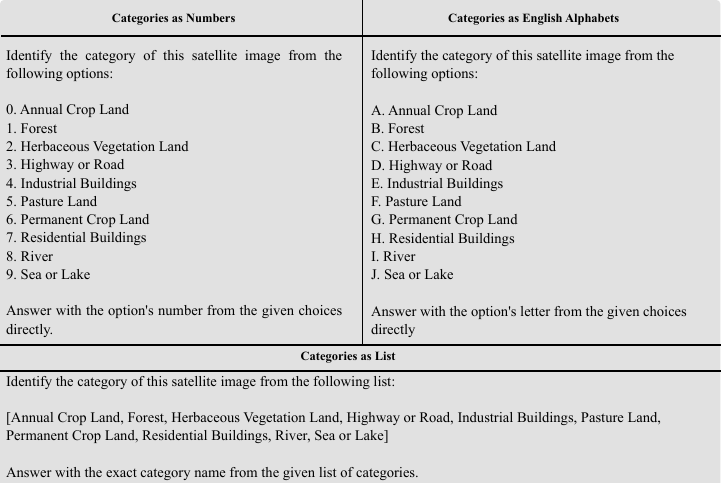}
    \caption{Example of different prompting options explored for LLAVA for EuroSAT~\cite{eurosat}.}
    \label{fig:llava_prompting}
\end{figure}

%% file: supp_sections/ensembling_descriptions.tex
\section{Ensembling Descriptions}
\label{sec:supp:ensembling_descriptions}
\input{supp_tables/ensembling_descriptions}

In the main manuscript (Table~3) we provide results by constructing the zero-shot classifier by ensembling the VLM prompts in two different ways:

\paragraph{Embedding Space:} The zero-shot classifier is constructed as the mean of the embeddings (from the text encoder of CLIP~\cite{clip}) from the different sources (\eg Mixtral~\cite{mixtral} or GPT~\cite{gpt3}) VLM prompts). 
\paragraph{Probability Space:} The zero-shot classifier is constructed as the mean of the probabilities (\eg from softmax) obtained by different VLM prompt sources (\eg Mixtral~\cite{mixtral} or GPT~\cite{gpt3}) for \method. 

In Table~3 (main manuscript) we observed different behaviors (in terms of the obtained results) from these two sources of ensembling. 
In theory, an ensemble over the probability space can also be obtained for the individual category-specific VLM prompts (from stage 2) of the \method. 
However, for datasets with a larger number of classes (\eg ImageNet~\cite{imagenet} with $1000$ classes), such an ensemble is prohibitively expensive  (as also noted in~\cite{clip}).
Nevertheless, for completeness, in Table~\ref{tab:ensemble_comparison}, we provide results for the two ensembling methods for datasets with a smaller number of classes.
From these results, we observe that the two different ensembling methods do not result in a huge deviation in performance. 
Note, to obtain all the \method results in all our experiments reported in the main paper, we always construct the zero-shot classifier as the mean of the embeddings from the VLM prompts for each category.


%% file: supp_tables/ensembling_descriptions.tex
\begin{table}[t!]
    \centering
                        \setlength{\tabcolsep}{10pt}

    \begin{tabular}{l|cccc|c}
\toprule
    \multicolumn{6}{c}{Ensemble in Embedding Space}\\
    \midrule
\midrule
         &  EuroSAT & Flowers & DTD & Resisc&Mean\\
        Top-1 (\%) - ViT-B/32 & 55.6 &73.9&50.8&64.0&61.1\\
\midrule
            \multicolumn{6}{c}{Ensemble in Probability Space}\\
\midrule
\midrule
         &  EuroSAT & Flowers & DTD & Resisc&Mean\\

        Top-1 (\%) - ViT-B/32 & 54.5&73.0&51.0&61.3&60.0\\
\bottomrule
\bottomrule
    \end{tabular}
    \caption{Comparison of constructing the zero-shot classifier by ensembling the GPT \method prompts over the embedding or probability space.}
    \label{tab:ensemble_comparison}
\end{table}

%% file: supp_sections/rebuttal_exp.tex
\begin{table}[t!]
\small
    \centering
    \begin{minipage}{0.4\textwidth}
    \centering
    \resizebox{1.2\textwidth}{!}{\begin{tabular}{cc|cc}
    \toprule
         CLIP&CLIP$+$MPVR&GEM&GEM$+$MPVR\\
         \midrule
         \midrule
          11.2&\textbf{15.0}&46.2&\textbf{51.3}\\
          \bottomrule
          \bottomrule
    \end{tabular}}
    \end{minipage}\hfill
    \begin{minipage}{0.4\textwidth}
    \centering
     \resizebox{0.9\textwidth}{!}{\begin{tabular}{l|cc}
     \toprule
         &  base&novel\\
         \midrule
         \midrule
          OVD&56.6&36.9\\
          OVD+MPVR&\textbf{57.1}&\textbf{40.6}\\
          \bottomrule
          \bottomrule
    \end{tabular}}
    \end{minipage}
    \caption{\textbf{Left:} Semantic Segmentation mIOU (CLIP ViT-B/16) on Pascal VOC. \textbf{Right:} Object Detection mAP@50 on MS-COCO.}
    \label{tab:supp:detection-segmentation-results}
\end{table}

\begin{figure}
    \centering
    \includegraphics[width=1\linewidth]{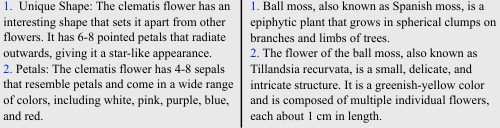}
    \caption{Qualitative Examples}
    \label{fig:supp:qualitative-examples}
\end{figure}

\begin{table}[t!]
    \begin{minipage}{1\textwidth}
    \tiny
        \setlength{\tabcolsep}{1.5pt}
    \centering
     \resizebox{0.65\textwidth}{!}{\begin{tabular}{l|cc}
     \toprule
         datasets&direct-replace&MPVR\\
         \midrule
         \midrule
          flowers&66.9&75.2\\
                    sun&63.4&67.0\\
          food&78.5&81.3\\
          eurosat&50.3&55.6\\
          \midrule
          mean&64.8&69.8\\
          \bottomrule
          \bottomrule
    \end{tabular}}
     \caption{Top-1 accuracy by directly replacing the task-specific information in the in-context prompts and the 2-stage MPVR.}
    \label{tab:supp:manual-replacement}
    \end{minipage}
\end{table}

\begin{table*}[t!]
    \centering
\resizebox{\textwidth}{!}{\begin{tabular}{lcccccccccc}
\toprule
 &\textbf{EuroSAT} & \textbf{INR} & \textbf{Flowers} & \textbf{INS} & \textbf{DTD} & \textbf{FGVCAircraft} & \textbf{Food} & \textbf{kinetics400} & \textbf{Caltech101} & \textbf{places365} \\
 \midrule 
 \midrule
 OpenCLIP&42.9 & 74.4 & 69.8 & 53.0 & 49.2 & 23.0 & 78.2 & 38.6 & 95.9 & 42.1 
\\
 MPVR (GPT)&57.6 & 77.6 & 74.3 & 54.9 & 61.7 & 26.0 & 78.7 & 42.3 & 94.8 & 43.7
\\
 \midrule
 SigLip&41.3&89.3&84.3&67.1&62.1&40.7&89.1&46.1&97.8&41.4\\
 MPVR (GPT)&46.4 & 90.3 & 88.7 & 68.3 & 66.7 & 45.9 & 88.5 & 48.4 & 97.2 & 43.6\\
 \midrule
 & \textbf{CUBS200} & \textbf{ImageNet} & \textbf{Cars} & \textbf{SUN397} & \textbf{ImageNetV2} & \textbf{CIFAR10} & \textbf{CIFAR100} & \textbf{OxfordPets} & \textbf{UCF101} & \textbf{RESISC45}\\
 \midrule
  OpenCLIP& 65.2 & 66.1 & 88.3 & 68.2 & 57.9 & 93.7 & 75.8 & 87.3 & 63.4 & 55.9\\
 MPVR (GPT)&67.0 & 67.0 & 88.2 & 69.6 & 59.0 & 93.9 & 75.8 & 91.4 & 66.9 & 66.6\\
 \midrule
 SigLip&65.5&75.7&90.7&69.6&68.4&92.5&70.9&93.2&70.8&60.3\\
 MPVR (GPT)&66.3 & 76.2 & 90.3 & 70.9 & 69.0 & 92.6 & 71.1 & 93.9 & 69.6 & 64.3\\
\bottomrule
\bottomrule
\end{tabular}}
\caption{Top-1 Accuracy (\%) for OpenCLIP (ViT-B/32) and SigLIP (ViT-B/16).}
\label{tab:supp:siglip_openclip_results}
\end{table*}

\section{Additional Insights and Experiments}
\label{sec:supp:additional-insights-experiments}
This section provides additional insights and experiments the reviewers requested during the review process. 
First, we examine the role of two-stage prompting in MPVR, then study the concerns of data leakage (due to LLM already knowing the downstream datasets) and also look into detail why adding the class information in the meta prompt hurt the MPVR performance, later provide a few qualitative examples and finally conclude with results for additional downstream tasks and comparison with OpenCLIP~\cite{openclip} and SigLIP~\cite{siglip} backbones.

\paragraph{Effect of 2-Stages in MPVR:} The role of in-context examples provided to the LLM is only to specify the desired output format to the LLM (\ie~a \texttt{python} code with category name placeholders).
To further analyze if `LLM merely replaces the corresponding part of in-context prompts', we manually replaced the downstream task specification (\eg~\texttt{texture} $\to$ \texttt{flower}, for Oxford Flowers dataset) in the in-context prompts (provided in the supplementary Fig. 1) and generated the VLM prompts directly from stage 2 (circumventing stage 1).
Results in Table~\ref{tab:supp:manual-replacement} 
show that our proposed meta-prompting allows for more diverse task-adaptive LLM knowledge extraction, not possible through simple heuristic replacement.

\paragraph{Class names in meta prompt and data leakage}
We manually verified the LLM queries generated from meta-prompts with and without additional class name information and found: 1) with class names, the generated queries are less diverse and focus on specifics, thus restricting the diversity of the VLM prompts; 2) some queries are not syntactically correct, resulting in less effective VLM prompts.
To test against leakage of dataset information, we build a new dataset dubbed as \texttt{Mixture Dataset}, by combining Flowers, Textures, and EuroSAT datasets. 
Our MPVR-generated VLM prompts improve the CLIP 0-shot accuracy from $46.7\%~\to~51.7\%$, suggesting that the performance gains with MPVR are not due to data leakage. 

\paragraph{Qualitative Examples}
In the two randomly chosen prompts from the best performing \textit{Clematis} (left) and worst performing \textit{Ball moss} (right) category in the Oxford Flowers dataset (Figure~\ref{fig:supp:qualitative-examples})
we observe that instead of describing the~\textit{Ball moss} flower in the UK, the prompt is about the Spanish moss of the same name.

\paragraph{Additional VLMs and downstream tasks} MPVR also scales to VLMs trained on different sources of data. 
We see an average improvement of $3.3\%$ ($64.4\% \to 67.8\%$) and $1.6\%$ ($70.8\% \to 72.4\%$) for \textbf{OpenCLIP} and \textbf{SigLIP}. 
Dataset-wise results are provided in Table~\ref{tab:supp:siglip_openclip_results}. 
Furthermore, in Table~\ref{tab:supp:detection-segmentation-results} we see that our MPVR improves upon vanilla CLIP and the training-free SOTA method~GEM~\cite{gem} for semantic segmentation, and also improves Open Vocabulary Object Detector OVD~\cite{ovd}. 

%% file: supp_sections/detailed_results.tex
\section{Detailed Results}
\label{sec:supp:detailed_results}
\input{supp_tables/embedding_ensembles_expanded_tables}

\input{supp_tables/softmax_ensembles_expanded}

\input{supp_tables/vit_b16_results}

\input{supp_tables/vit_l_14_results}

\input{supp_tables/metaclip_b32}

\input{supp_tables/metaclip_b16}

\input{supp_tables/metaclip_l14}

For completeness, here we provide dataset-wise results for two experiments in the main manuscript: ensembling different text sources (Table~3) and mean results (over $20$ datasets) for different backbones, listed in Table~2. 
To this end, in Table~\ref{tab:supp:ensemble_emb_space} and Table~\ref{tab:supp:ensemble_prob_space} we provide the dataset-wise results by mixing different text sources and ensembling these sources either in the embedding space or the probability space. 
In Tables~\ref{tab:supp:vit_b16_results}~\&~\ref{tab:supp:vit-l14} we provide the dataset-wise results for ViT-B/16 and ViT-L/14 from CLIP~\cite{clip}.
Furthermore, in Tables~\ref{tab:supp:metaclip_b32_results},~\ref{tab:supp:metaclip_b16_results}~\&~\ref{tab:supp:metaclip_l14_results} we list the detailed results for $3$ different backbones (ViT-B/32, ViT-B/16 and ViT-L/14) from MetaCLIP~\cite{metaclip}. 
The detailed (dataset-wise) results also highlight that our \method performs favorably on most datasets when compared to the state-of-the-art methods. 

%% file: supp_tables/embedding_ensembles_expanded_tables.tex
\begin{table*}[t!]
    \centering
\resizebox{\textwidth}{!}{\begin{tabular}{lcccccccccc}
\toprule
\multicolumn{11}{c}{\textbf{ViT-B/32}}\\
\midrule
 & \textbf{ImageNet} & \textbf{ImageNetv2} & \textbf{C10} & \textbf{C100} & \textbf{Caltech101} & \textbf{Flowers} & \textbf{Stanford Cars} & \textbf{Cubs} & \textbf{Pets} & \textbf{DTD} \\
\midrule
\midrule
GPT+TEMP & {45.9} & {40.2} & {\underline{87.6}} & {59.0} & {88.1} & {72.6} & {\underline{40.8}} & {54.6} & {86.2} & {48.3} \\
MIXTRAL+TEMP & {\underline{46.7}} & {\underline{41.2}} & {\underline{87.6}} & {59.0} & {88.8} & {74.4} & {40.0} & {55.4} & {86.5} & {47.5} \\
GPT+MIXTRAL & {\bfseries{64.9}} & {\bfseries{57.4}} & {\bfseries{89.8}} & {\bfseries{66.0}} & {\bfseries{92.8}} & {\bfseries{76.0}} & {\bfseries{59.2}} & {\bfseries{55.8}} & {\bfseries{88.6}} & {\bfseries{51.1}} \\
GPT+MIXTRAL+TEMP & {46.1} & {40.8} & {86.7} & {\underline{60.7}} & {\underline{89.9}} & {\bfseries{76.0}} & {40.7} & {\underline{55.7}} & {\underline{87.9}} & {\underline{49.2}} \\
\midrule
& \textbf{Food101} & \textbf{Aircraft} & \textbf{Places365} & \textbf{SUN397} & \textbf{UCF101} & \textbf{K400} & \textbf{IN-R} & \textbf{IN-S} & \textbf{EuroSAT} & \textbf{Resisc45} \\
\midrule
\midrule
GPT+TEMP & {80.1} & {20.9} & {42.3} & {66.1} & {62.8} & {36.6} & {58.8} & {29.3} & {\bfseries{56.6}} & {62.2} \\
MIXTRAL+TEMP & {\underline{80.8}} & {\underline{22.2}} & {41.9} & {66.0} & {59.8} & {35.7} & {61.1} & {15.5} & {50.6} & {61.6} \\
GPT+MIXTRAL & {\bfseries{81.1}} & {\bfseries{22.3}} & {\bfseries{42.7}} & {\bfseries{67.1}} & {\bfseries{67.1}} & {\bfseries{43.4}} & {\bfseries{70.3}} & {\bfseries{44.0}} & {\underline{56.4}} & {\bfseries{65.0}} \\
GPT+MIXTRAL+TEMP & {80.2} & {21.1} & {\underline{42.6}} & {\underline{66.2}} & {\underline{63.0}} & {\underline{37.7}} & {\underline{61.9}} & {\underline{29.5}} & {55.3} & {\underline{63.3}} \\

\midrule
\midrule

\multicolumn{11}{c}{\textbf{ViT-B/16}}\\
\midrule
 & \textbf{ImageNet} & \textbf{ImageNetv2} & \textbf{C10} & \textbf{C100} & \textbf{Caltech101} & \textbf{Flowers} & \textbf{Stanford Cars} & \textbf{Cubs} & \textbf{Pets} & \textbf{DTD} \\
\midrule
\midrule

GPT+TEMP & {49.9} & {44.8} & {84.6} & {63.7} & {89.2} & {76.1} & {45.8} & {58.6} & {87.6} & {53.2} \\
MIXTRAL+TEMP & {\underline{50.5}} & {\underline{45.5}} & {\underline{86.9}} & {62.6} & {89.8} & {78.3} & {44.5} & {59.7} & {89.6} & {50.0} \\
GPT+MIXTRAL & {\bfseries{69.7}} & {\bfseries{63.4}} & {\bfseries{91.2}} & {\bfseries{69.6}} & {\bfseries{94.4}} & {\bfseries{79.2}} & {\bfseries{65.6}} & {\underline{59.8}} & {\bfseries{90.7}} & {\bfseries{55.4}} \\
GPT+MIXTRAL+TEMP & {50.2} & {45.1} & {85.8} & {\underline{64.3}} & {\underline{91.5}} & {\underline{78.9}} & {\underline{46.4}} & {\bfseries{60.1}} & {\underline{90.1}} & {\underline{53.4}} \\
\midrule
& \textbf{Food101} & \textbf{Aircraft} & \textbf{Places365} & \textbf{SUN397} & \textbf{UCF101} & \textbf{K400} & \textbf{IN-R} & \textbf{IN-S} & \textbf{EuroSAT} & \textbf{Resisc45} \\
\midrule
\midrule
GPT+TEMP & {85.8} & {27.8} & {43.1} & {67.9} & {67.0} & {40.8} & {64.2} & {\underline{36.0}} & {58.5} & {65.4} \\
MIXTRAL+TEMP & {\underline{86.2}} & {\bfseries{29.3}} & {42.7} & {\underline{68.2}} & {65.3} & {39.8} & {66.4} & {17.1} & {55.0} & {63.8} \\
GPT+MIXTRAL & {\bfseries{86.5}} & {\underline{28.3}} & {\bfseries{43.6}} & {\bfseries{68.8}} & {\bfseries{70.1}} & {\bfseries{48.0}} & {\bfseries{78.4}} & {\bfseries{50.6}} & {\bfseries{60.2}} & {\bfseries{67.2}} \\
GPT+MIXTRAL+TEMP & {86.0} & {28.1} & {\bfseries{43.6}} & {68.1} & {\underline{67.4}} & {\underline{42.1}} & {\underline{67.8}} & {\underline{36.0}} & {\underline{59.1}} & {\underline{66.1}} \\

\midrule
\midrule

\multicolumn{11}{c}{\textbf{ViT-L/14}}\\
\midrule
 & \textbf{ImageNet} & \textbf{ImageNetv2} & \textbf{C10} & \textbf{C100} & \textbf{Caltech101} & \textbf{Flowers} & \textbf{Stanford Cars} & \textbf{Cubs} & \textbf{Pets} & \textbf{DTD} \\
\midrule
\midrule
GPT+TEMP & {62.2} & {\underline{57.3}} & {90.1} & {75.5} & {94.3} & {81.9} & {58.9} & {64.8} & {92.9} & {62.0} \\
MIXTRAL+TEMP & {\underline{62.4}} & {56.9} & {\underline{93.3}} & {76.7} & {93.6} & {82.0} & {54.9} & {66.6} & {92.2} & {60.4} \\
GPT+MIXTRAL & {\bfseries{76.9}} & {\bfseries{71.0}} & {\bfseries{96.2}} & {\bfseries{79.4}} & {\bfseries{95.5}} & {\bfseries{83.9}} & {\bfseries{78.1}} & {\bfseries{67.3}} & {\bfseries{93.8}} & {\bfseries{62.9}} \\
GPT+MIXTRAL+TEMP & {\underline{62.4}} & {\underline{57.3}} & {91.9} & {\underline{76.8}} & {\underline{94.7}} & {\underline{82.5}} & {\underline{59.2}} & {\underline{67.2}} & {\underline{93.2}} & {\underline{62.8}} \\
\midrule
& \textbf{Food101} & \textbf{Aircraft} & \textbf{Places365} & \textbf{SUN397} & \textbf{UCF101} & \textbf{K400} & \textbf{IN-R} & \textbf{IN-S} & \textbf{EuroSAT} & \textbf{Resisc45} \\
\midrule
\midrule
GPT+TEMP & {91.2} & {33.1} & {43.4} & {72.6} & {73.2} & {50.7} & {81.4} & {\underline{49.5}} & {\bfseries{67.2}} & {70.0} \\
MIXTRAL+TEMP & {91.3} & {\bfseries{36.1}} & {42.7} & {72.3} & {73.7} & {49.8} & {82.9} & {28.2} & {60.4} & {69.1} \\
GPT+MIXTRAL & {\bfseries{91.6}} & {\underline{34.3}} & {\bfseries{43.8}} & {\bfseries{72.9}} & {\bfseries{77.4}} & {\bfseries{55.5}} & {\bfseries{88.6}} & {\bfseries{61.2}} & {\underline{65.6}} & {\bfseries{71.5}} \\
GPT+MIXTRAL+TEMP & {\underline{91.4}} & {33.5} & {\bfseries{43.8}} & {\underline{72.7}} & {\underline{74.4}} & {\underline{51.7}} & {\underline{83.5}} & {49.4} & {65.4} & {\underline{70.2}} \\
\bottomrule
\bottomrule
\end{tabular}}
\caption{Top-1 accuracy (\%) while ensembling different text sources in the embedding space. Here, the zero-shot classifier is constructed by taking the mean of the embeddings from the different individual text sources. 
}
\label{tab:supp:ensemble_emb_space}
\end{table*}

%% file: supp_tables/softmax_ensembles_expanded.tex
\begin{table*}[t!]
    \centering
\resizebox{\textwidth}{!}{\begin{tabular}{lcccccccccc}
\toprule
\multicolumn{11}{c}{\textbf{ViT-B/32}}\\
\midrule
 & \textbf{ImageNet} & \textbf{ImageNetv2} & \textbf{C10} & \textbf{C100} & \textbf{Caltech101} & \textbf{Flowers} & \textbf{Stanford Cars} & \textbf{Cubs} & \textbf{Pets} & \textbf{DTD} \\
\midrule
\midrule
GPT+TEMP & {\bfseries{64.8}} & {\bfseries{57.2}} & {\bfseries{89.9}} & {\bfseries{66.3}} & {92.7} & {73.7} & {\bfseries{59.4}} & {\underline{55.8}} & {88.3} & {\underline{51.3}} \\
MIXTRAL+TEMP & {\underline{63.8}} & {\underline{56.3}} & {89.5} & {65.5} & {\bfseries{93.0}} & {75.2} & {\underline{58.2}} & {55.5} & {88.4} & {50.3} \\
GPT+MIXTRAL & {62.9} & {55.9} & {89.7} & {\underline{66.1}} & {92.8} & {\underline{76.1}} & {52.4} & {\bfseries{55.9}} & {\underline{88.8}} & {51.2} \\
GPT+MIXTRAL+TEMP & {62.9} & {55.8} & {\underline{89.8}} & {\underline{66.1}} & {\underline{92.9}} & {\bfseries{76.2}} & {52.4} & {\underline{55.8}} & {\bfseries{89.0}} & {\bfseries{51.4}} \\
\midrule
& \textbf{Food101} & \textbf{Aircraft} & \textbf{Places365} & \textbf{SUN397} & \textbf{UCF101} & \textbf{K400} & \textbf{IN-R} & \textbf{IN-S} & \textbf{EuroSAT} & \textbf{Resisc45} \\
\midrule
\midrule
GPT+TEMP & {\underline{81.1}} & {21.9} & {42.1} & {\bfseries{67.0}} & {\bfseries{68.0}} & {\bfseries{43.7}} & {70.1} & {\bfseries{44.2}} & {55.0} & {63.9} \\
MIXTRAL+TEMP & {\bfseries{81.2}} & {\bfseries{22.3}} & {42.1} & {\underline{66.5}} & {66.4} & {42.2} & {70.1} & {42.7} & {53.3} & {64.5} \\
GPT+MIXTRAL & {79.2} & {22.2} & {\underline{42.6}} & {66.2} & {\underline{67.2}} & {\underline{43.4}} & {\bfseries{70.3}} & {43.8} & {\bfseries{56.4}} & {\bfseries{65.0}} \\
GPT+MIXTRAL+TEMP & {79.2} & {\bfseries{22.3}} & {\bfseries{42.7}} & {66.2} & {\underline{67.2}} & {\underline{43.4}} & {\bfseries{70.3}} & {\underline{43.9}} & {\underline{56.1}} & {\bfseries{65.0}} \\

\midrule
\midrule

\multicolumn{11}{c}{\textbf{ViT-B/16}}\\
\midrule
 & \textbf{ImageNet} & \textbf{ImageNetv2} & \textbf{C10} & \textbf{C100} & \textbf{Caltech101} & \textbf{Flowers} & \textbf{Stanford Cars} & \textbf{Cubs} & \textbf{Pets} & \textbf{DTD} \\
\midrule
\midrule

GPT+TEMP & {\bfseries{69.8}} & {\bfseries{63.2}} & {90.8} & {\underline{69.7}} & {94.0} & {76.8} & {\bfseries{65.4}} & {58.8} & {89.8} & {\bfseries{56.6}} \\
MIXTRAL+TEMP & {\underline{68.8}} & {\underline{62.2}} & {\underline{91.1}} & {69.2} & {94.0} & {78.2} & {\underline{62.2}} & {\bfseries{60.3}} & {90.4} & {54.1} \\
GPT+MIXTRAL & {67.7} & {61.6} & {\underline{91.1}} & {\underline{69.7}} & {\bfseries{94.4}} & {\underline{79.4}} & {57.0} & {\underline{60.1}} & {\underline{91.0}} & {55.6} \\
GPT+MIXTRAL+TEMP & {67.7} & {61.5} & {\bfseries{91.2}} & {\bfseries{69.8}} & {\bfseries{94.4}} & {\bfseries{79.7}} & {57.0} & {\underline{60.1}} & {\bfseries{91.1}} & {\underline{55.9}} \\
\midrule
& \textbf{Food101} & \textbf{Aircraft} & \textbf{Places365} & \textbf{SUN397} & \textbf{UCF101} & \textbf{K400} & \textbf{IN-R} & \textbf{IN-S} & \textbf{EuroSAT} & \textbf{Resisc45} \\
\midrule
\midrule
GPT+TEMP & {\underline{86.4}} & {27.5} & {43.0} & {\bfseries{68.9}} & {\bfseries{70.7}} & {\bfseries{48.0}} & {78.3} & {50.7} & {59.1} & {66.1} \\
MIXTRAL+TEMP & {\bfseries{86.6}} & {\bfseries{29.8}} & {42.6} & {\underline{68.7}} & {68.8} & {46.9} & {\underline{78.4}} & {49.7} & {59.0} & {66.7} \\
GPT+MIXTRAL & {84.7} & {\underline{28.1}} & {\bfseries{43.5}} & {68.4} & {\underline{70.0}} & {\bfseries{48.0}} & {\underline{78.4}} & {\underline{50.8}} & {\bfseries{60.3}} & {\bfseries{67.1}} \\
GPT+MIXTRAL+TEMP & {84.6} & {28.0} & {\bfseries{43.5}} & {68.4} & {69.8} & {47.9} & {\bfseries{78.5}} & {\bfseries{50.9}} & {\underline{60.0}} & {\underline{67.0}} \\

\midrule
\midrule

\multicolumn{11}{c}{\textbf{ViT-L/14}}\\
\midrule
 & \textbf{ImageNet} & \textbf{ImageNetv2} & \textbf{C10} & \textbf{C100} & \textbf{Caltech101} & \textbf{Flowers} & \textbf{Stanford Cars} & \textbf{Cubs} & \textbf{Pets} & \textbf{DTD} \\
\midrule
\midrule
GPT+TEMP & {\bfseries{76.8}} & {\bfseries{70.9}} & {95.8} & {79.1} & {\bfseries{96.2}} & {83.7} & {\bfseries{78.4}} & {65.2} & {93.6} & {\bfseries{62.9}} \\
MIXTRAL+TEMP & {\underline{75.9}} & {69.6} & {96.1} & {79.3} & {95.3} & {83.6} & {\underline{70.9}} & {\bfseries{67.5}} & {93.0} & {61.6} \\
GPT+MIXTRAL & {75.2} & {\underline{69.7}} & {\bfseries{96.2}} & {\bfseries{79.5}} & {\underline{95.8}} & {\bfseries{84.4}} & {65.7} & {67.3} & {\bfseries{93.9}} & {62.7} \\
GPT+MIXTRAL+TEMP & {75.2} & {\underline{69.7}} & {\bfseries{96.2}} & {\bfseries{79.5}} & {95.7} & {\underline{84.3}} & {65.8} & {\underline{67.4}} & {\bfseries{93.9}} & {\bfseries{62.9}} \\
\midrule
& \textbf{Food101} & \textbf{Aircraft} & \textbf{Places365} & \textbf{SUN397} & \textbf{UCF101} & \textbf{K400} & \textbf{IN-R} & \textbf{IN-S} & \textbf{EuroSAT} & \textbf{Resisc45} \\
\midrule
\midrule
GPT+TEMP & {\bfseries{91.5}} & {34.3} & {43.5} & {\bfseries{72.9}} & {\bfseries{77.9}} & {\bfseries{55.7}} & {88.5} & {\underline{61.1}} & {\bfseries{67.6}} & {71.0} \\
MIXTRAL+TEMP & {\underline{91.4}} & {\bfseries{37.5}} & {42.5} & {\underline{72.4}} & {75.9} & {54.6} & {\bfseries{88.6}} & {60.0} & {61.9} & {71.1} \\
GPT+MIXTRAL & {90.6} & {\underline{35.6}} & {\bfseries{43.8}} & {72.1} & {\underline{77.5}} & {\underline{55.6}} & {\bfseries{88.6}} & {\underline{61.1}} & {\underline{65.2}} & {\bfseries{71.7}} \\
GPT+MIXTRAL+TEMP & {90.6} & {35.5} & {\bfseries{43.8}} & {72.2} & {\underline{77.5}} & {\underline{55.6}} & {\bfseries{88.6}} & {\bfseries{61.2}} & {65.1} & {\underline{71.6}} \\
\bottomrule
\bottomrule
\end{tabular}}
\caption{Top-1 accuracy (\%) while ensembling different text sources in the probability space. Here, the zero-shot classifier is constructed by taking the mean of the softmax probabilities from the different individual classifiers. 
}
\label{tab:supp:ensemble_prob_space}
\end{table*}

%% file: supp_tables/vit_b16_results.tex
\begin{table*}[t!]
    \centering

\resizebox{\textwidth}{!}{\begin{tabular}{lcccccccccc}
\toprule
 & \textbf{ImageNet} & \textbf{ImageNetv2} & \textbf{C10} & \textbf{C100} & \textbf{Caltech101} & \textbf{Flowers} & \textbf{Stanford Cars} & \textbf{Cubs} & \textbf{Pets} & \textbf{DTD} \\
\midrule
\midrule
S-TEMP & {66.7} & {60.9} & {90.1} & {68.4} & {93.3} & {67.5} & {\underline{65.5}} & {55.1} & {88.2} & {43.3} \\
DS-TEMP & {68.3} & {61.9} & {\underline{90.8}} & {68.2} & {92.9} & {70.7} & {\bfseries{66.2}} & {56.1} & {89.1} & {43.2} \\
CUPL & {\bfseries{69.7}} & {\bfseries{63.4}} & {90.3} & {69.0} & {\underline{94.4}} & {70.9} & {60.0} & {56.0} & {\bfseries{91.2}} & {53.3} \\
D-CLIP & {68.6} & {62.2} & {89.6} & {68.4} & {\bfseries{94.5}} & {72.1} & {63.7} & {56.7} & {\underline{90.3}} & {42.8} \\
Waffle & {68.3} & {62.3} & {\underline{90.8}} & {68.8} & {93.7} & {72.2} & {64.0} & {57.0} & {89.2} & {41.9} \\
Waffle+Con & {68.3} & {62.3} & \underline{90.8} & {68.8} & {90.7} & {69.0} & {63.9} & {56.5} & {89.4} & {42.7} \\
Waffle+Con+GPT & {68.3} & {62.3} & \underline{90.8} & {68.8} & {\underline{94.4}} & {72.3} & {63.8} & {56.8} & {89.7} & {42.8} \\
\midrule
MPVR (Mixtral) & {68.8} & {62.2} & {\bfseries{91.1}} & {\underline{69.1}} & {94.2} & {\bfseries{78.4}} & {62.2} & {\bfseries{60.4}} & {\underline{90.3}} & {\underline{53.7}} \\
MPVR (GPT) & {\bfseries{69.7}} & {\textbf{63.4}} & {\underline{90.8}} & {\bfseries{69.5}} & {94.1} & {\underline{76.9}} & {65.4} & {\underline{59.0}} & {89.9} & {\bfseries{56.1}} \\
\midrule
& \textbf{Food101} & \textbf{Aircraft} & \textbf{Places365} & \textbf{SUN397} & \textbf{UCF101} & \textbf{K400} & \textbf{IN-R} & \textbf{IN-S} & \textbf{EuroSAT} & \textbf{Resisc45} \\
\midrule
\midrule
S-TEMP & {85.2} & {23.8} & {39.3} & {62.5} & {65.1} & {43.7} & {74.0} & {46.2} & {42.3} & {56.5} \\
DS-TEMP & {85.9} & {24.3} & {40.9} & {65.3} & {68.5} & {\underline{47.4}} & {77.7} & {48.8} & {48.9} & {60.1} \\
CUPL & {86.1} & {26.6} & {\_} & {\bfseries{69.0}} & {\underline{68.9}} & {46.0} & {\_} & {\_} & {\_} & {\underline{66.2}} \\
D-CLIP & {86.1} & {24.0} & {42.0} & {66.1} & {67.5} & {45.2} & {76.5} & {48.9} & {58.5} & {64.8} \\
Waffle & {\bfseries{86.9}} & {24.9} & {42.0} & {65.4} & {67.1} & {46.1} & {77.0} & {49.1} & {49.6} & {64.8} \\
Waffle+Con & {86.5} & {24.2} & {39.8} & {62.9} & {66.5} & {45.1} & {76.3} & {48.2} & {48.1} & {61.7} \\
Waffle+Con+GPT & {\underline{86.7}} & {24.9} & {42.4} & {66.4} & {68.4} & {46.0} & {77.0} & {49.5} & {55.6} & {65.2} \\
\midrule
MPVR (Mixtral) & {86.6} & {\bfseries{29.9}} & {\underline{42.7}} & {\underline{68.9}} & {\underline{68.9}} & {46.9} & {\bfseries{78.4}} & {\underline{49.7}} & {\underline{59.2}} & {\bfseries{66.7}} \\
MPVR (GPT) & {86.4} & {\underline{28.0}} & {\bfseries{43.1}} & {68.8} & {\bfseries{70.9}} & {\bfseries{48.0}} & {\underline{78.2}} & {\bfseries{50.6}} & {\bfseries{59.6}} & {\underline{66.2}} \\
\bottomrule
\bottomrule
\end{tabular}}
\caption{Top-1 accuracy (\%) for 20 datasets obtained by employing the ViT-B/16 backbone from OpenAI CLIP~\cite{clip}. 
S-TEMP refers to the results obtained by using the
default template (\texttt{a photo of a <class name>}), while DS-TEMP refers to the results
obtained by using the ensemble of dataset-specific prompts. An empty placeholder
for CUPL [34] indicates that the respective baseline did not provide the handcrafted
prompts for the dataset. For Waffle~\cite{waffle}, mean results from 7 random runs are reported,
following the original publication.
}
\label{tab:supp:vit_b16_results}
\end{table*}

%% file: supp_tables/vit_l_14_results.tex
\begin{table*}[t!]
    \centering

\resizebox{\textwidth}{!}{\begin{tabular}{lcccccccccc}
\toprule
 & \textbf{ImageNet} & \textbf{ImageNetv2} & \textbf{C10} & \textbf{C100} & \textbf{Caltech101} & \textbf{Flowers} & \textbf{Stanford Cars} & \textbf{Cubs} & \textbf{Pets} & \textbf{DTD} \\
\midrule
\midrule
S-TEMP & {73.5} & {67.8} & {95.2} & {77.2} & {94.3} & {76.2} & {76.9} & {62.1} & {93.1} & {52.5} \\
DS-TEMP & {75.5} & {69.9} & {95.7} & {78.3} & {93.7} & {79.5} & {\underline{78.1}} & {61.8} & {93.5} & {54.8} \\
CUPL & {\underline{76.7}} & {\underline{70.8}} & {95.8} & {78.6} & {96.1} & {79.6} & {64.2} & {60.3} & {\bfseries{94.3}} & {61.1} \\
D-CLIP & {75.1} & {69.0} & {95.2} & {78.4} & {\bfseries{97.0}} & {79.5} & {75.1} & {61.7} & {93.0} & {56.1} \\
Waffle & {75.1} & {68.9} & {\underline{96.0}} & {78.4} & {96.2} & {78.3} & {76.5} & {62.3} & {93.2} & {55.3} \\
Waffle+Con & {75.1} & {68.9} & {\underline{96.0}} & {78.4} & {93.9} & {77.3} & {76.7} & {63.1} & {93.4} & {53.7} \\
Waffle+Con+GPT & {75.1} & {68.9} & {\underline{96.0}} & {78.4} & {\underline{96.9}} & {79.0} & {75.9} & {62.0} & {93.1} & {56.1} \\
\midrule
MPVR (Mixtral) & {75.9} & {69.6} & {\bfseries{96.1}} & {\bfseries{79.3}} & {95.4} & {\bfseries{83.8}} & {70.6} & {\bfseries{67.7}} & {93.1} & {\underline{61.6}} \\
MPVR (GPT) & {\bfseries{76.8}} & {\bfseries{70.9}} & {\underline{96.0}} & {\underline{79.2}} & {96.1} & {\underline{83.6}} & {\bfseries{78.3}} & {\underline{65.5}} & {\underline{93.7}} & {\bfseries{62.9}}\\
\midrule
& \textbf{Food101} & \textbf{Aircraft} & \textbf{Places365} & \textbf{SUN397} & \textbf{UCF101} & \textbf{K400} & \textbf{IN-R} & \textbf{IN-S} & \textbf{EuroSAT} & \textbf{Resisc45} \\
\midrule
\midrule
S-TEMP & {90.3} & {30.0} & {40.1} & {67.6} & {73.8} & {51.3} & {85.4} & {58.3} & {55.1} & {63.2} \\
DS-TEMP & {90.9} & {31.8} & {41.2} & {69.0} & {76.2} & {\underline{55.0}} & {87.8} & {59.8} & {63.2} & {68.0} \\
CUPL & {91.4} & {\underline{35.1}} & {\_} & {\underline{72.8}} & {75.8} & {54.4} & {\_} & {\_} & {\_} & {\bfseries{71.8}} \\
D-CLIP & {91.1} & {31.8} & {42.3} & {69.6} & {76.2} & {52.5} & {86.8} & {59.0} & {54.6} & {70.7} \\
Waffle & {\bfseries{91.5}} & {32.5} & {42.6} & {69.4} & {76.0} & {53.4} & {87.4} & {59.1} & {50.4} & {\underline{71.4}} \\
Waffle+Con & {91.2} & {31.3} & {41.1} & {66.2} & {74.2} & {52.0} & {86.2} & {58.6} & {44.2} & {66.7} \\
Waffle+Con+GPT & {91.4} & {32.1} & {\underline{42.9}} & {70.1} & {\underline{76.4}} & {53.5} & {87.3} & {59.3} & {53.7} & {71.1} \\
\midrule
MPVR (Mixtral) & {91.4} & {\bfseries{37.6}} & {42.5} & {72.5} & {75.8} & {54.6} & {\bfseries{88.5}} & {\underline{60.0}} & {62.2} & {71.2} \\
MPVR (GPT) & {\bfseries{91.5}} & {34.4} & {\bfseries{43.5}} & {\bfseries{73.0}} & {\bfseries{78.1}} & {\bfseries{55.7}} & {\underline{88.4}} & {\bfseries{61.0}} & {\bfseries{67.3}} & {71.1} \\
\bottomrule
\bottomrule
\end{tabular}}
\caption{Top-1 accuracy (\%) for 20 datasets obtained by employing the ViT-L/14 backbone from OpenAI CLIP~\cite{clip}. 
}
\label{tab:supp:vit-l14}
\end{table*}

%% file: supp_tables/metaclip_b32.tex
\begin{table*}[t!]
    \centering

\resizebox{\textwidth}{!}{\begin{tabular}{lcccccccccc}
\toprule
 & \textbf{ImageNet} & \textbf{ImageNetv2} & \textbf{C10} & \textbf{C100} & \textbf{Caltech101} & \textbf{Flowers} & \textbf{Stanford Cars} & \textbf{Cubs} & \textbf{Pets} & \textbf{DTD} \\
\midrule
\midrule
S-TEMP & {64.1} & {56.3} & {91.2} & {66.8} & {\bfseries{95.5}} & {69.8} & {\underline{71.7}} & {61.8} & {86.9} & {47.6} \\
DS-TEMP & {65.6} & {\underline{57.5}} & {\underline{91.3}} & {\bfseries{70.2}} & {93.8} & {71.3} & {\bfseries{72.1}} & {62.5} & {\bfseries{88.7}} & {51.8} \\
CUPL & {\bfseries{66.0}} & {\underline{57.5}} & {90.3} & {68.4} & {\bfseries{95.5}} & {68.3} & {61.3} & {61.5} & {88.5} & {58.4} \\
D-CLIP & {64.0} & {55.0} & {90.9} & {68.4} & {94.9} & {67.6} & {66.6} & {62.1} & {87.9} & {50.0} \\
Waffle & {63.5} & {55.5} & {90.9} & {67.2} & {93.9} & {69.7} & {68.8} & {61.7} & {88.6} & {50.0} \\
Waffle+Con & {63.5} & {55.5} & {90.9} & {67.2} & {88.2} & {68.6} & {69.1} & {61.8} & {\bfseries{88.7}} & {48.5} \\
Waffle+Con+GPT & {63.5} & {55.5} & {90.9} & {67.2} & {94.8} & {68.7} & {68.1} & {62.1} & {88.1} & {51.0} \\
\midrule
MPVR (Mixtral) & {64.8} & {57.4} & {91.2} & {68.9} & {94.3} & {\bfseries{78.4}} & {68.3} & {\bfseries{65.2}} & {88.1} & {\bfseries{61.5}} \\
MPVR (GPT) & {\bfseries{66.0}} & {\bfseries{57.6}} & {\bfseries{91.4}} & {\underline{69.2}} & {94.5} & {\underline{74.8}} & {71.2} & {\underline{64.6}} & {88.0} & {\bfseries{61.5}} \\
\midrule
& \textbf{Food101} & \textbf{Aircraft} & \textbf{Places365} & \textbf{SUN397} & \textbf{UCF101} & \textbf{K400} & \textbf{IN-R} & \textbf{IN-S} & \textbf{EuroSAT} & \textbf{Resisc45} \\
\midrule
\midrule
S-TEMP & {76.7} & {24.3} & {39.6} & {64.8} & {64.4} & {36.9} & {71.5} & {52.3} & {49.4} & {56.4} \\
DS-TEMP & {\bfseries{77.3}} & {26.9} & {40.1} & {65.3} & {\underline{66.1}} & {39.1} & {74.8} & {\underline{53.9}} & {50.4} & {60.6} \\
CUPL & {77.0} & {\underline{32.3}} & {\_} & {\bfseries{67.7}} & {64.2} & {39.3} & {\_} & {\_} & {\_} & {\underline{63.9}} \\
D-CLIP & {76.7} & {25.3} & {42.0} & {64.3} & {65.6} & {37.6} & {73.2} & {52.3} & {49.0} & {62.4} \\
Waffle & {\underline{77.2}} & {26.0} & {42.1} & {65.8} & {64.1} & {38.1} & {73.9} & {52.9} & {42.3} & {\bfseries{64.4}} \\
Waffle+Con & {77.1} & {25.4} & {41.4} & {66.0} & {63.5} & {37.4} & {72.8} & {52.8} & {37.8} & {63.6} \\
Waffle+Con+GPT & {\underline{77.2}} & {25.7} & {\bfseries{42.4}} & {65.6} & {65.8} & {38.4} & {73.8} & {52.9} & {46.7} & {63.4} \\
\midrule
MPVR (Mixtral) & {77.0} & {\bfseries{35.4}} & {41.4} & {\underline{67.3}} & {65.4} & {\underline{39.9}} & {\bfseries{75.7}} & {53.1} & {\underline{56.3}} & {61.4} \\
MPVR (GPT) & {77.1} & {31.8} & {\bfseries{42.4}} & {65.8} & {\bfseries{67.3}} & {\bfseries{40.6}} & {\underline{75.6}} & {\bfseries{54.1}} & {\bfseries{58.7}} & {63.6} \\
\bottomrule
\bottomrule
\end{tabular}}
\caption{Top-1 accuracy (\%) for 20 datasets obtained by employing the ViT-B/32 backbone from MetaCLIP~\cite{metaclip}.}
\label{tab:supp:metaclip_b32_results}
\end{table*}

%% file: supp_tables/metaclip_b16.tex
\begin{table*}[t!]
    \centering

\resizebox{\textwidth}{!}{\begin{tabular}{lcccccccccc}
\toprule
 & \textbf{ImageNet} & \textbf{ImageNetv2} & \textbf{C10} & \textbf{C100} & \textbf{Caltech101} & \textbf{Flowers} & \textbf{Stanford Cars} & \textbf{Cubs} & \textbf{Pets} & \textbf{DTD} \\
\midrule
\midrule
S-TEMP & {70.0} & {61.8} & {\underline{89.9}} & {64.9} & {\underline{95.7}} & {71.7} & {74.7} & {69.5} & {88.5} & {53.0} \\
DS-TEMP & {70.8} & {\underline{62.6}} & {\bfseries{90.1}} & {\underline{66.5}} & {95.6} & {73.8} & {\underline{75.8}} & {69.7} & {90.5} & {56.3} \\
CUPL & {\underline{70.9}} & {62.5} & {89.2} & {65.5} & {95.5} & {70.8} & {0.5} & {68.9} & {89.8} & {62.2} \\
D-CLIP & {69.0} & {60.7} & {88.6} & {64.6} & {\underline{95.7}} & {72.7} & {71.9} & {68.4} & {90.1} & {53.5} \\
Waffle & {69.1} & {61.0} & {87.9} & {64.9} & {95.0} & {73.3} & {73.1} & {68.2} & {\underline{90.8}} & {53.5} \\
Waffle+Con & {69.1} & {61.0} & {87.9} & {64.9} & {94.1} & {72.1} & {72.3} & {68.5} & {\bfseries{91.0}} & {52.5} \\
Waffle+Con+GPT & {69.1} & {61.0} & {87.9} & {64.9} & {\bfseries{95.8}} & {72.9} & {73.0} & {68.1} & {90.7} & {55.1} \\
\midrule
MPVR (Mixtral) & {69.8} & {62.0} & {89.8} & {65.6} & {95.5} & {\bfseries{80.6}} & {74.0} & {\underline{71.2}} & {90.4} & {\underline{64.1}} \\
MPVR (GPT) & {\bfseries{71.2}} & {\bfseries{62.9}} & {89.8} & {\bfseries{66.6}} & {94.8} & {\underline{75.9}} & {\bfseries{75.9}} & {\bfseries{71.4}} & {89.9} & {\bfseries{64.4}} \\
\midrule
& \textbf{Food101} & \textbf{Aircraft} & \textbf{Places365} & \textbf{SUN397} & \textbf{UCF101} & \textbf{K400} & \textbf{IN-R} & \textbf{IN-S} & \textbf{EuroSAT} & \textbf{Resisc45} \\
\midrule
\midrule
S-TEMP & {83.8} & {26.3} & {41.6} & {68.8} & {67.0} & {39.9} & {80.1} & {56.5} & {50.9} & {63.5} \\
DS-TEMP & {\bfseries{84.1}} & {28.3} & {41.7} & {68.4} & {\bfseries{69.0}} & {43.2} & {81.8} & {\bfseries{58.7}} & {55.2} & {63.9} \\
CUPL & {\underline{84.0}} & {\underline{34.4}} & {\_} & {\bfseries{69.5}} & {66.6} & {43.3} & {\_} & {\_} & {\_} & {67.0} \\
D-CLIP & {83.7} & {30.1} & {42.5} & {66.8} & {67.2} & {41.6} & {79.5} & {57.1} & {56.1} & {67.3} \\
Waffle & {83.9} & {30.5} & {42.3} & {67.7} & {68.4} & {42.4} & {80.0} & {56.9} & {53.3} & {67.8} \\
Waffle+Con & {83.9} & {30.2} & {41.5} & {68.2} & {66.3} & {41.7} & {79.4} & {57.5} & {49.7} & {67.8} \\
Waffle+Con+GPT & {\underline{84.0}} & {30.4} & {\underline{42.8}} & {68.0} & {68.5} & {42.4} & {80.1} & {57.2} & {55.9} & {\underline{68.3}} \\
\midrule
MPVR (Mixtral) & {\underline{84.0}} & {\bfseries{37.8}} & {41.4} & {\underline{69.4}} & {67.9} & {\underline{44.2}} & {\bfseries{82.2}} & {57.2} & {\bfseries{59.7}} & {67.0} \\
MPVR (GPT) & {83.6} & {34.0} & {\bfseries{43.0}} & {\underline{69.4}} & {\underline{68.8}} & {\bfseries{44.9}} & {\underline{82.1}} & {\underline{58.2}} & {\underline{57.8}} & {\bfseries{69.1}} \\
\bottomrule
\bottomrule
\end{tabular}}
\caption{Top-1 accuracy (\%) for 20 datasets obtained by employing the ViT-B/16 backbone from MetaCLIP~\cite{metaclip}.}
\label{tab:supp:metaclip_b16_results}
\end{table*}

%% file: supp_tables/metaclip_l14.tex
\begin{table*}[t!]
    \centering

\resizebox{\textwidth}{!}{\begin{tabular}{lcccccccccc}
\toprule
 & \textbf{ImageNet} & \textbf{ImageNetv2} & \textbf{C10} & \textbf{C100} & \textbf{Caltech101} & \textbf{Flowers} & \textbf{Stanford Cars} & \textbf{Cubs} & \textbf{Pets} & \textbf{DTD} \\
\midrule
\midrule
S-TEMP & {75.1} & {68.5} & {94.9} & {74.4} & {96.8} & {76.7} & {\underline{84.5}} & {76.0} & {88.7} & {58.8} \\
DS-TEMP & {76.2} & {\underline{69.9}} & {95.7} & {\bfseries{77.4}} & {96.3} & {77.4} & {\bfseries{84.9}} & {75.2} & {\bfseries{93.7}} & {60.5} \\
CUPL & {\underline{76.5}} & {\underline{69.9}} & {95.0} & {76.3} & {\underline{97.0}} & {75.8} & {81.1} & {74.4} & {92.7} & {64.5} \\
D-CLIP & {74.4} & {67.8} & {95.7} & {75.9} & {\underline{97.0}} & {76.7} & {82.9} & {74.7} & {93.0} & {58.0} \\
Waffle & {74.3} & {67.9} & {95.6} & {76.6} & {96.2} & {78.3} & {83.0} & {74.5} & {92.9} & {59.6} \\
Waffle+Con & {74.3} & {67.9} & {95.6} & {76.6} & {95.3} & {78.6} & {83.8} & {75.0} & {\underline{93.2}} & {57.0} \\
Waffle+Con+GPT & {74.3} & {67.9} & {95.6} & {76.6} & {\bfseries{97.4}} & {77.5} & {83.2} & {74.5} & {93.0} & {60.2} \\
\midrule
MPVR (Mixtral) & {75.5} & {68.6} & {\bfseries{95.9}} & {76.5} & {96.6} & {\bfseries{85.5}} & {82.2} & {\bfseries{77.9}} & {92.6} & {\bfseries{67.3}} \\
MPVR (GPT) & {\bfseries{76.6}} & {\bfseries{70.1}} & {95.1} & {76.0} & {96.0} & {\underline{84.9}} & {83.7} & {\underline{77.6}} & {93.0} & {\underline{65.8}} \\
\midrule
& \textbf{Food101} & \textbf{Aircraft} & \textbf{Places365} & \textbf{SUN397} & \textbf{UCF101} & \textbf{K400} & \textbf{IN-R} & \textbf{IN-S} & \textbf{EuroSAT} & \textbf{Resisc45} \\
\midrule
\midrule
S-TEMP & {88.6} & {35.6} & {42.2} & {72.1} & {75.2} & {48.6} & {87.7} & {63.9} & {49.7} & {61.6} \\
DS-TEMP & {88.5} & {40.0} & {42.0} & {72.0} & {75.9} & {51.0} & {88.9} & {\underline{65.3}} & {56.8} & {69.1} \\
CUPL & {\underline{89.0}} & {41.2} & {\_} & {{71.9}} & {75.0} & {51.1} & {\_} & {\_} & {\_} & {{71.2}} \\
D-CLIP & {88.4} & {39.5} & {43.5} & {71.1} & {75.9} & {49.5} & {87.7} & {64.2} & {\bfseries{61.3}} & {67.9} \\
Waffle & {88.7} & {39.0} & {43.3} & {71.7} & {75.9} & {49.8} & {87.5} & {64.0} & {59.6} & {69.5} \\
Waffle+Con & {88.9} & {38.8} & {41.4} & {71.4} & {75.1} & {49.2} & {87.2} & {64.1} & {59.3} & {65.2} \\
Waffle+Con+GPT & {88.7} & {39.7} & {{43.0}} & {72.3} & {\underline{76.3}} & {50.2} & {87.9} & {64.3} & {\bfseries{61.3}} & {69.0} \\
\midrule
MPVR (Mixtral) & {\bfseries{89.1}} & {\bfseries{49.5}} & {40.2} & {\bfseries{73.1}} & {74.8} & {\underline{51.8}} & {\bfseries{89.4}} & {65.1} & {56.3} & {70.5} \\
MPVR (GPT) & {88.8} & {\underline{46.7}} & {\bfseries{43.8}} & {72.5} & {\bfseries{77.5}} & {\bfseries{52.3}} & {\underline{89.2}} & {\bfseries{65.5}} & {58.0} & {\textbf{72.8}} \\
\bottomrule
\bottomrule
\end{tabular}}
\caption{Top-1 accuracy (\%) for 20 datasets obtained by employing the ViT-L/14 backbone from MetaCLIP~\cite{metaclip}.}

\label{tab:supp:metaclip_l14_results}
\end{table*}